\pdfoutput=1

\documentclass[11pt]{article}

\usepackage{ACL2023}

\usepackage{times}
\usepackage{latexsym}

\usepackage[T1]{fontenc}

\usepackage[utf8]{inputenc}

\usepackage{microtype}

\usepackage{inconsolata}

\usepackage{enumitem}
\usepackage{graphicx}

\usepackage{color}
\usepackage{tikz}
\usepackage[edges]{forest}
\definecolor{hidden-draw}{RGB}{205, 44, 36}
\definecolor{hidden-blue}{RGB}{194,232,247}
\definecolor{hidden-orange}{RGB}{243,202,120}
\definecolor{hidden-yellow}{RGB}{242,244,193}
\definecolor{tree-level-1}{RGB}{245,20,85}
\definecolor{tree-level-2}{RGB}{246,86,118}
\definecolor{tree-level-3}{RGB}{248,177,193}
\definecolor{tree-leaf}{RGB}{176,230,198}

\definecolor{Self}{RGB}{255,0,128}
\definecolor{Ensemble}{RGB}{0,127,255}
\definecolor{Iterative}{RGB}{153,51,255}

\definecolor{exemplar1}{RGB}{136,98,148}
\definecolor{exemplar2}{RGB}{148,210,242}
\definecolor{knowledge1}{RGB}{249,219,152}
\definecolor{knowledge2}{RGB}{255,245,220}

\newcommand{\textq}[1]{{{#1}}}
\newcommand{\textr}[1]{{{#1}}}
\newcommand{\texta}[1]{{{#1}}}

\newcommand{\p}[1]{{\flushleft \textbf{#1.}}}
\newcommand{\ps}[1]{{\flushleft \textbf{#1?}}}

\title{Towards Reasoning in Large Language Models: A Survey}

\author{Jie Huang $\quad$ Kevin Chen-Chuan Chang \\
Department of Computer Science, University of Illinois at Urbana-Champaign \\
 \texttt{\{jeffhj, kcchang\}@illinois.edu}
}

\begin{document}
\maketitle
\begin{abstract}
Reasoning is a fundamental aspect of human intelligence that plays a crucial role in activities such as problem solving, decision making, and critical thinking. In recent years, large language models (LLMs) have made significant progress in natural language processing, and there is observation that these models may exhibit reasoning abilities when they are sufficiently large. However, it is not yet clear to what extent LLMs are capable of reasoning. This paper provides a comprehensive overview of the current state of knowledge on reasoning in LLMs, including techniques for improving and eliciting reasoning in these models, methods and benchmarks for evaluating reasoning abilities, findings and implications of previous research in this field, and suggestions on future directions. Our aim is to provide a detailed and up-to-date review of this topic and stimulate meaningful discussion and future work.\footnote{Paperlist can be found at \url{https://github.com/jeffhj/LM-reasoning}.}
\end{abstract}

\section{Introduction}

Reasoning is a cognitive process that involves using evidence, arguments, and logic to arrive at conclusions or make judgments. It plays a central role in many intellectual activities, such as problem solving, decision making, and critical thinking. The study of reasoning is important in fields like psychology~\citep{wason1972psychology}, philosophy~\citep{passmore1961philosophical}, and computer science~\citep{huth2004logic}, as it helps individuals make decisions, solve problems, and think critically.

Recently, large language models (LLMs)  \citep[][\textit{inter alia}]{brown2020language,chowdhery2022palm,chung2022scaling,openai2022chatgpt} such as ChatGPT have made significant advancements in natural language processing and related fields. It has been shown that these models exhibit emergent behaviors, including the ability to ``reason'', when they are large enough \citep{wei2022emergent}. 
For example, by providing the models with ``\textit{chain of thoughts}'', i.e., reasoning exemplars, or a simple prompt ``\textit{Let’s think step by step}'', these models are able to answer questions with explicit reasoning steps \citep{wei2022chain,kojima2022large}, 
e.g., ``all whales are mammals, all mammals have kidneys; therefore, all whales have kidneys.''
This has sparked considerable interest in the community since reasoning ability is a hallmark of human intelligence that is frequently considered missed in current artificial intelligence systems \cite{marcus2020next,russin2020deep,mitchell2021abstraction,bommasani2021opportunities}. 

However, despite the strong performance of LLMs on certain reasoning tasks, 
it remains unclear whether LLMs are actually reasoning and to what extent they are capable of reasoning.
For example, \citet{kojima2022large} claim that ``LLMs are decent zero-shot reasoners (p.~1)'', while \citet{valmeekam2022large} conclude that ``LLMs are still far from achieving acceptable performance on common planning/reasoning tasks which pose no issues for humans to do (p.~2).''
This limitation is also stated by \citet{wei2022chain}: 
\begin{itemize}[label={},leftmargin=5mm,topsep=5pt]
    \item ``we qualify that although chain of thought emulates the thought processes of human reasoners, this does not answer whether the neural network is actually \textit{reasoning} (p.~9).''
\end{itemize}

Therefore, in this paper, we aim to provide a comprehensive overview and engage in an insightful discussion on the current state of knowledge on this fast-evolving topic. We initiate our exploration with a clarification of the concept of reasoning (\S \ref{sec:definition}). Subsequently, we turn our attention to the techniques for enhancing/eliciting reasoning in LLMs (\S \ref{sec:method}), the methods and benchmarks for evaluating reasoning in LLMs (\S \ref{sec:evaluation}), and the key findings and implications in this field (\S \ref{sec:findings}). Finally, we reflect on and discuss the current state of the field (\S \ref{sec:discussion}).

\tikzstyle{my-box}=[
    rectangle,
    draw=hidden-draw,
    rounded corners,
    text opacity=1,
    minimum height=1.5em,
    minimum width=5em,
    inner sep=2pt,
    align=center,
    fill opacity=.5,
]
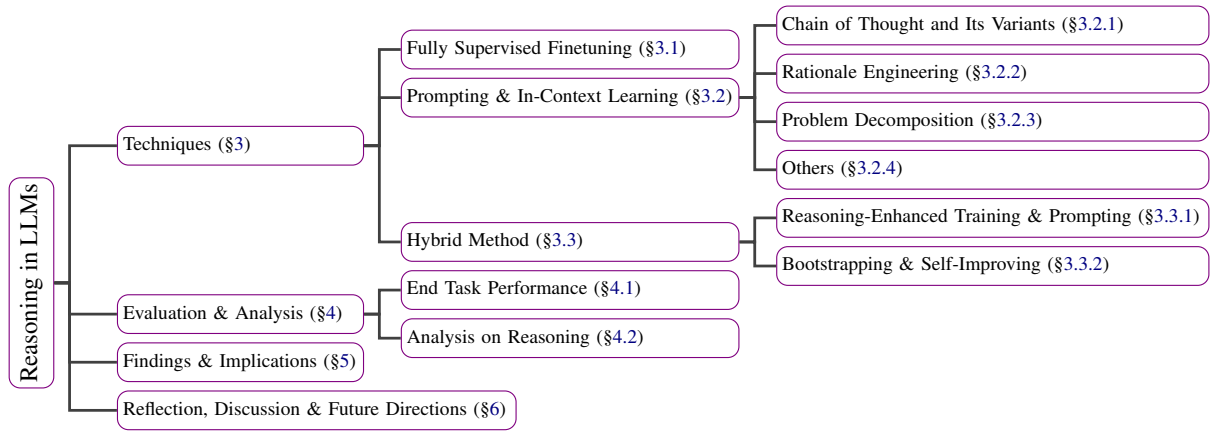
\begin{figure*}[tp]
    \centering
    \resizebox{\linewidth}{!}{
        \begin{forest}
            forked edges,
            for tree={
                grow=east,
                reversed=true,
                anchor=base west,
                parent anchor=east,
                child anchor=west,
                base=left,
                font=\small,
                rectangle,
                draw=violet,
                rounded corners,
                align=left,
                minimum width=4em,
                edge+={darkgray, line width=1pt},
                s sep=3pt,
                inner xsep=2pt,
                inner ysep=3pt,
                ver/.style={rotate=90, child anchor=north, parent anchor=south, anchor=center},
            },
            where level=1{text width=7.7em,font=\scriptsize,}{},
            where level=2{text width=10.7em,font=\scriptsize,}{},
            where level=3{text width=13.8em,font=\scriptsize,}{},
            [
                Reasoning in LLMs, ver
                [
                    Techniques (\S \ref{sec:method})
                    [
                        Fully Supervised Finetuning (\S \ref{sec:fully-sup})
                    ]
                    [
                        Prompting \& In-Context Learning (\S \ref{sec:prompting})
                        [
                            Chain of Thought and Its Variants (\S\ref{sec:cot})
                        ]
                        [
                            Rationale Engineering (\S\ref{sec:rationale-eng})
                        ]  
                        [
                            Problem Decomposition  (\S\ref{sec:prob-decom})
                        ]  
                        [
                            Others (\S\ref{sec:others_prompting})
                        ]
                    ]
                    [
                        Hybrid Method (\S \ref{sec:hybrid})
                        [
                            Reasoning-Enhanced Training \& Prompting (\S \ref{sec:trainig_and_prompting})
                        ]
                        [
                            Bootstrapping \& Self-Improving (\S \ref{sec:selfimprove})
                        ]
                    ]
                ]
                [
                    Evaluation \& Analysis (\S \ref{sec:evaluation})
                    [
                        End Task Performance (\S \ref{sec:endtask})
                    ]
                    [
                        Analysis on Reasoning (\S\ref{sec:analysis})
                    ]
                ]
                [
                    Findings \& Implications (\S\ref{sec:findings})
                ]
                [
                    Reflection{,} Discussion \& Future Directions (\S\ref{sec:discussion}), text width=12.7em
                ]
            ]
        \end{forest}
    }
    \caption{The structure of the paper.}
    \label{fig:taxo}
\end{figure*}

\section{What is Reasoning?}
\label{sec:definition}

Reasoning is the process of thinking about something in a logical and systematic way, using evidence and past experiences to reach a conclusion or make a decision \citep{wason1972psychology,wason1968reasoning,galotti1989approaches,fagin2004reasoning,mchugh2018reasoning}. Reasoning involves making inferences, evaluating arguments, and drawing logical conclusions based on available information. 
Although ``reasoning'' is a term that is commonly used in literature and daily life, it is also an abstract concept that can refer to many things.
To help the reader better understand this concept, we summarize several main categories of reasoning that are commonly recognized:

\p{Deductive reasoning} Deductive reasoning is a type of reasoning in which a conclusion is drawn based on the truth of the premises. In deductive reasoning, the conclusion must necessarily follow from the premises, meaning that if the premises are true, the conclusion must also be true. For example:
\begin{itemize}[noitemsep,topsep=3pt]
    \item Premise: All mammals have kidneys.
    \item Premise: All whales are mammals.
    \item Conclusion: All whales have kidneys.
\end{itemize}

\p{Inductive reasoning} Inductive reasoning is a type of reasoning in which a conclusion is drawn based on observations or evidence. The conclusion is likely to be true based on the available evidence, but it is not necessarily certain. For example:
\begin{itemize}[noitemsep,topsep=3pt]
    \item Observation: Every time we see a creature with wings, it is a bird.
    \item Observation: We see a creature with wings.
    \item Conclusion: The creature is likely to be a bird.
\end{itemize}

\p{Abductive reasoning} Abductive reasoning is a type of reasoning in which a conclusion is drawn based on the best explanation for a given set of observations. The conclusion is the most likely explanation based on the available evidence, but it is not necessarily certain. For example:
\begin{itemize}[noitemsep,topsep=3pt]
    \item Observation: The car cannot start and there is a puddle of liquid under the engine.
    \item Conclusion: The most likely explanation is that the car has a leak in the radiator.
\end{itemize}

\noindent Other types of reasoning include \textit{analogical reasoning}, which involves making comparisons between two or more things in order to make inferences or arrive at conclusions; \textit{causal reasoning}, which involves identifying and understanding the causes and effects of events or phenomena; and \textit{probabilistic reasoning}, which involves making decisions or arriving at conclusions based on the likelihood or probability of certain outcomes.

\p{Formal Reasoning vs Informal Reasoning} 
\textit{Formal reasoning} is a systematic and logical process that follows a set of rules and principles, often used in mathematics and logic. \textit{Informal reasoning} is a less structured approach that relies on intuition, experience, and common sense to draw conclusions and solve problems, and is often used in everyday life. Formal reasoning is more structured and reliable, while informal reasoning is more adaptable and open-ended, but may also be less reliable. We refer the reader to \citet{galotti1989approaches,bronkhorst2020logical} for a detailed distinction between them.

\p{Reasoning in Language Models}
The concept of reasoning in language models has been around for some time, but there is not a clear definition of what it entails.
In the literature, the term ``reasoning'' is often used to refer to informal reasoning, although it is not always explicitly stated that it is informal~\citep[][\textit{inter alia}]{cobbe2021training,wei2022chain}.
Different forms of reasoning may be used depending on the task, benchmark, or method being used, e.g., deductive reasoning~\citep[][\textit{inter alia}]{cobbe2021training,creswell2022selection,han2022human}, inductive reasoning~\citep[][\textit{inter alia}]{yang2022language,misra2022property} or abductive reasoning~\citep[][\textit{inter alia}]{wiegreffe2021reframing,lampinen2022can,jung2022maieutic}.
In this paper, we encompass various forms of reasoning, with a particular focus on ``informal deductive reasoning'' in large language models since it is a widely used form in which the conclusion is guaranteed to be true as long as the premises are true.

\section{Towards Reasoning in Large Language Models}

\label{sec:method}

Reasoning, particularly multi-step reasoning, is often seen as a weakness in language models and other NLP models \citep{bommasani2021opportunities,rae2021scaling,valmeekam2022large}. 
Recent research has suggested that reasoning ability may emerge in language models at a certain scale, such as models with over 100 billion parameters \citep{wei2022emergent,wei2022chain,cobbe2021training}. In this paper, we follow \citet{wei2022emergent} in considering reasoning as an ability that is rarely present in small-scale models like GPT-2~\citep{radford2019language} and BERT~\citep{kenton2019bert}, and therefore focus on techniques applicable to improving or eliciting ``reasoning''\footnote{It is important to note that the term ``reasoning'' in this paper does not necessarily imply that LLMs are truly capable of reasoning or that they are able to reason in the same way that humans do. We will discuss this issue in more detail in~\S \ref{sec:discussion}.} in LLMs such as GPT-3~\citep{brown2020language} and PaLM~\citep{chowdhery2022palm}. 

\subsection{Fully Supervised Finetuning}
\label{sec:fully-sup}

Before discussing reasoning in large language models, it is worth mentioning there is research working on eliciting/improving reasoning in small language models through \textit{fully supervised finetuning} on specific datasets.
For example, \citet{rajani2019explain} finetune a pretrained GPT model~\citep{radford2018improving} to generate rationales that explain model predictions with the built CoS-E dataset, and find that models trained with explanations perform better on commonsense question answering tasks~\citep{talmor2019commonsenseqa}.
\citet{talmor2020leap} train RoBERTa~\citep{liu2019roberta} to perform reasoning/inference based on both implicit pre-trained knowledge and explicit free-text statements. 
\citet{dan2021@measuring} finetune pretrained language models to solve competition mathematics problems by generating full step-by-step solutions, though the accuracy is relatively low.
\citet{nye2021show} train language models to do multi-step reasoning for program synthesis/execution by generating ``scratchpads'', i.e., intermediate computations, before producing the final answers.  We refer the reader to \citet{helwe2021reasoning,bhargava2022commonsense}'s survey for more studies in this line.

There are two major limitations of fully supervised finetuning. First, it requires a dataset containing explicit reasoning, which can be difficult and time-consuming to create. Additionally, the model is only trained on a specific dataset, which limits its application to a specific domain and may result in the model relying on artifacts in the training data rather than actual reasoning to make predictions.

\subsection{Prompting \& In-Context Learning}
\label{sec:prompting}

\begin{figure*}[tp!]
\centerline{\includegraphics[width=0.9\linewidth]{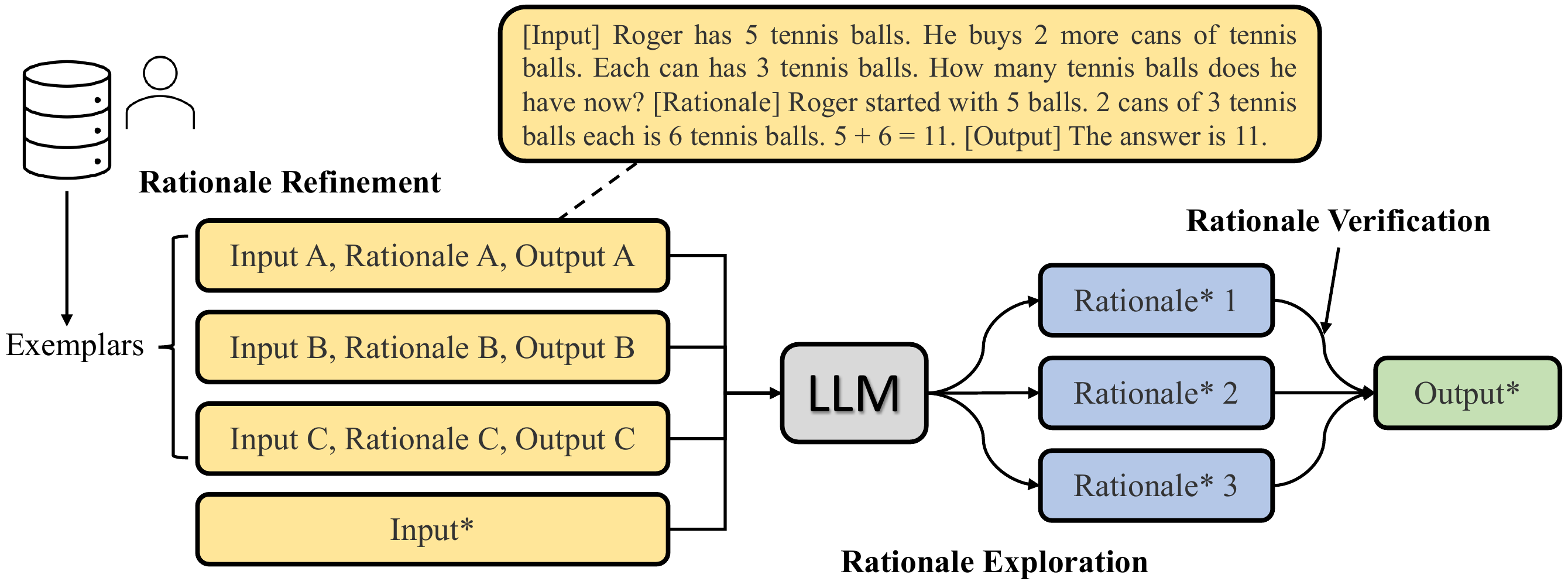}}
\caption{An illustration of \textit{Chain-of-Thought Prompting} and \textit{Rationale Engineering}, where asterisk (*) denotes the target problem to be solved.}
\label{fig:RE}
\end{figure*}

Large language models such as GPT-3 \citep{brown2020language} have demonstrated remarkable few-shot performance across a variety of tasks through in-context learning.
These models can be prompted with a question and a few $\langle$input, output$\rangle$ exemplars to potentially solve a problem through ``reasoning'', either implicitly or explicitly.
However, research has shown that these models still fall short when it comes to tasks that require multiple steps of reasoning to solve \citep{bommasani2021opportunities,rae2021scaling,valmeekam2022large}. This may be due to a lack of exploration into the full capabilities of these models, as recent studies have suggested.

\subsubsection{Chain of Thought and Its Variants}
\label{sec:cot}

To encourage LLMs to engage in reasoning rather than simply providing answers directly, we may guide LLMs to generate ``reasoning'' explicitly. 
One approach for doing this is \textit{chain-of-thought prompting}, proposed by \citet{wei2022chain}.
This approach involves providing a few examples of ``chain of thought'' (CoT), which are intermediate natural language reasoning steps, in the prompt to LLMs (Figure~\ref{fig:RE}).
Specifically, in CoT prompting, $\langle$input, output$\rangle$ demonstrations are replaced with $\langle$\textq{input}, \textr{\textit{chain of thought}}, \texta{output}$\rangle$ triples, e.g., ``\textq{[input] Roger has 5 tennis balls. He buys 2 more cans of tennis balls. Each can has 3 tennis balls. How many tennis balls does he have now?} \textr{[\textit{chain of thought}] Roger started with 5 balls. 2 cans of 3 tennis balls each is 6 tennis balls. 5 + 6 = 11.} \texta{[output] The answer is 11.}'' 
In this way, given a target question, the model learns to generate explicit rationale before producing the final answer. 
Experimental results show that this simple idea can improve LLMs' few-shot performance on arithmetic, symbolic, and commonsense reasoning tasks, sometimes to a striking degree.

There are several variants of chain-of-thought prompting that have been proposed in the literature, in a different form or to solve a specific problem.

\textit{Different Form:}
\citet{kojima2022large} introduce \textit{Zero-shot-CoT}, in which LLMs are simply prompted with the phrase ``Let's think step by step'' after the input, in order to elicit reasoning without the need for few-shot demonstrations. 
\citet{madaan2022language,gao2022pal,chen2022program} find that LLMs trained with code, e.g., Codex~\cite{chen2021evaluating}, can achieve better performance on reasoning tasks by framing reasoning as code generation.
\citet{Wang2022iteratively} propose to iteratively prompt chain of thought.
\citet{he2022rethinking} attempt to retrieve external knowledge in CoT to improve faithfulness of reasoning.

\textit{Specific Problem/Setting:}
Before chain of thought, \citet{nye2021show} also try to use intermediate computations, named ``scratchpads'', to improve language models' reasoning performance in both finetuning and few-shot regimes, with a particular focus on programs.
\citet{shi2022language} attempt to solve multilingual reasoning tasks with CoT in the native language, CoT in English (regardless of the problem language), and CoT in English (with the problem translated to English).
\citet{chen2022large} apply CoT to table-based reasoning, finding that LLMs can achieve strong performance on table tasks with only one exemplar. 
\citet{prystawski2022psychologically} demonstrate that CoT can improve LLMs' performance on paraphrase selection for metaphors. \citet{lu2022learn} apply chain of thought to solve multimodal science questions. 

\subsubsection{Rationale Engineering}
\label{sec:rationale-eng}

The original version of chain-of-thought prompting, proposed by \citet{wei2022chain}, relies on manually crafted examples of intermediate reasoning steps and applies greedy decoding in the generation.
\textit{Rationale engineering} aims to more effectively elicit or utilize reasoning in LLMs.
This can be achieved through \textit{rationale refinement}, which involves creating more effective examples of reasoning steps, or through \textit{rationale exploration} and \textit{rationale verification}, which involve exploring and verifying the rationales produced by LLMs.
A summary of raltionale engineering is illustrated in Figure~\ref{fig:RE}.

\p{Rationale refinement} 
The choice of exemplars can significantly affect the few-shot performance of LLMs, as demonstrated in research such as \citet{liu2022makes}, which also appears in chain-of-thought prompting.
\textit{Rationale refinement} aims to create and refine rationale examples that are better able to elicit reasoning in LLMs.
\citet{fu2022complexity} propose \textit{complexity-based prompting} to create rationales with more reasoning steps. Their experiments show that the performance of LLMs improves with the increased rationale complexity. 
Similarly, \citet{zhou2022teaching} propose \textit{algorithmic prompting}, which suggests that providing more thorough examples of solutions can help improve reasoning performance on some simple math calculations.
\citet{zhang2022automatic} design \textit{Auto-CoT} to automatically construct exemplars by partitioning questions from a given dataset into clusters and then using Zero-Shot-CoT~\citep{kojima2022large} to generate the rationale for a representative question from each cluster. The analysis shows that making exemplars diverse is important in prompting LLMs to produce better rationales.

\p{Rationale exploration} In addition to providing better exemplars, we can allow LLMs to fully explore various ways of reasoning to improve their performance on reasoning tasks, named \textit{rationale exploration}.
Based on the idea that complex problems often admit multiple ways of thinking that can lead to their unique correct answer,
\citet{wang2022self} present a decoding strategy called \textit{self-consistency} to improve upon the traditional greedy decoding used in chain-of-thought prompting. This strategy involves sampling a diverse set of rationales, rather than just the greedy one, and selecting the most consistent answer by marginalizing out the sampled rationales. The idea is also used in \citet{fu2022complexity} to vote over the top complex rationales.
To further improve performance, \citet{li2022advance} suggest providing different demonstrations for each question by sampling exemplars from an exemplar base, in order to increase the diversity of the sampled rationales.

\p{Rationale verification} 
Ensuring that the rationales produced by LLMs are valid is critical, as incorrect rationales can lead to incorrect final predictions~\citep{ye2022unreliability}. To address this issue, the process of \textit{rationale verification} aims to verify whether the rationales produced by LLMs lead to the correct final answers. \citet{cobbe2021training} propose augmenting LLMs with a trained verifier that assigns a score to each rationale and solution generated by the LLM, selecting the highest-ranked solution as the final answer when solving math word problems.
\citet{li2022advance} also use this technique to guide rationale selection, in conjunction with the process of rationale exploration.
Different from the above methods that train an external verifier to verify the rationales, \citet{weng2022large} suggest using LLMs themselves as the verifiers. 

\subsubsection{Problem Decomposition}
\label{sec:prob-decom}

Chain-of-thought prompting, while effective for eliciting reasoning in LLMs, can struggle with complex tasks, e.g., tasks that require compositional generalization \citep{lake2018generalization,keysers2019measuring}.
To solve a complex problem, it is helpful to first break it down into smaller, more manageable subproblems. By solving each of these subproblems, we can effectively solve the complex problem. This technique is called \textit{problem decomposition} or \textit{divide and conquer}~\citep{talmor2018web,min2019multi,perez2020unsupervised}. 

Based on this idea, \citet{zhou2022least} propose \textit{least-to-most prompting}, which consists of two steps: decomposing the complex problem into subproblems and solving these subproblems in a specific order, with each subproblem being facilitated by the answers obtained from previously solved subproblems.
As follow-up work, \citet{drozdov2022compositional} introduce \textit{dynamic least-to-most prompting}, which is designed to solve more realistic semantic parsing problems by decomposing the problems with prompting-based syntactic parsing and dynamically selecting exemplars based on the decomposition.
In addition, \citet{khot2022decomposed} design \textit{decomposed prompting}, which breaks down a complex problem into subproblems that can be handled by a shared library of prompting-based LLMs, each specialized in a particular subproblem.
Furthermore, \citet{dua2022successive} develop \textit{successive prompting}, which iteratively decomposes a complex problem into a simple problem, with the next subproblem prediction having access to the answers to the previous subproblems. 
While the above methods decompose or solve compositional questions with multiple forward passes, \citet{press2022measuring} suggest decomposing and solving the input question in one forward pass using CoT prompting.
Overall, these techniques show promise for helping LLMs to solve complex tasks by decomposing the problem into more manageable subproblems.

\subsubsection{Others}
\label{sec:others_prompting}

There are other techniques that have been developed to facilitate reasoning in LLMs for specific tasks or settings.
For instance,
\citet{creswell2022selection,creswell2022faithful} introduce a \textit{selection-inference} framework that uses LLMs as modules to select and infer reasoning steps from a set of facts that culminate in the final answer.
\citet{kazemi2022lambada} suggest using backward chaining, i.e., from goal to the set of facts that support it, instead of forward chaining like \citet{creswell2022selection,creswell2022faithful}.
In addition,
\citet{jung2022maieutic} propose a method for solving binary questions by prompting LLMs abductively and recursively to rationalize each option.
\citet{zhou2022reflection} design a technique for performing numerical reasoning on complex numbers by replacing the complex numbers with simple numbers to produce simpler expressions, and then using these expressions to perform calculations on the complex numbers. 
There are also efforts to distill reasoning from LLMs into smaller models, such as the work by \citet{li2022explanations,shridhar2022distilling,magister2022@teaching}.
Finally, we refer the reader to \citet{dohan2022language}'s position paper on \textit{language model cascade}, which presents a unifying framework for understanding chain-of-thought prompting and research in this line.

\subsection{Hybrid Method}
\label{sec:hybrid}

While ``prompting'' techniques can help elicit or better utilize reasoning in large language models to solve reasoning tasks, they do not actually improve the reasoning capabilities of the LLMs themselves, as the parameters of the models remain unchanged. In contrast, the ``hybrid approach'' aims to simultaneously improve the reasoning capabilities of LLMs and make better use of these models in order to solve complex problems. This approach involves both enhancing the reasoning capabilities of the LLMs and using techniques such as prompting to effectively utilize these capabilities.

\subsubsection{Reasoning-Enhanced Training and Prompting}
\label{sec:trainig_and_prompting}

One approach to improving the reasoning capabilities of LLMs is to pretrain or finetune the models on datasets that include ``reasoning''.
\citet{lewkowycz2022solving,taylor2022galactica} find that LLMs trained on datasets containing scientific and mathematical data can achieve better performance on reasoning tasks like quantitative reasoning problems when using CoT prompting\footnote{This may also be true for models trained with code~\citep{chen2021evaluating,fu_khot_peng_2022b}.}.
\citet{pi2022reasoning} show that continually pretraining with SQL data can boost the performance of language models, e.g., T5~\citep{raffel2020exploring}, on natural language reasoning such as numerical reasoning and logical reasoning.
Furthermore, \citet{chung2022scaling} develop Flan models by finetuning PaLM~\citep{chowdhery2022palm} and T5~\citep{raffel2020exploring} with 1.8k finetuning tasks, including CoT data, and find that CoT data are critical to keeping reasoning abilities.
Similarly, \citet{yu2022alert} finetune OPT~\citep{zhang2022opt} on 10 reasoning datasets and observe that it can improve some reasoning capabilities of LLMs.
\citet{anil2022exploring} study the length generalization abilities of LLMs, i.e., whether LLMs learned with short problem instances can generalize to long ones. They discover that the combination of few-shot scratchpad (or chain of thought) finetuning and scratchpad prompting results in a significant improvement in LLMs' ability to generalize to longer problems, while this phenomenon is not observed in the standard fully supervised finetuning paradigm.

\subsubsection{Bootstrapping \& Self-Improving}
\label{sec:selfimprove}

Instead of finetuning LLMs on pre-built datasets that include reasoning, 
there are studies that have explored the idea of using LLMs to self-improve their reasoning abilities through a process known as bootstrapping.
One example of this is the \textit{Self-Taught Reasoner (STaR)} introduced by 
\citet{zelikman2022star}, in which a LLM is trained and refined on its own output iteratively. Specifically, with CoT prompting, the model first generates initial rationales. And then, the model is finetuned on rationales that lead to correct answers. This process can be repeated, with each iteration resulting in an improved model that can generate better training data, which in turn leads to further improvements.
As a follow-up to this work, \citet{huang2022large2} show that LLMs are able to self-improve their reasoning abilities without the need for supervised data by leveraging the self-consistency of reasoning~\citep{wang2022self}.

\section{Measuring Reasoning in Large Language Models}
\label{sec:evaluation}

 We summarize methods and benchmarks for evaluating reasoning abilities of LLMs in this section.

\subsection{End Task Performance}
\label{sec:endtask}

One way to measure reasoning abilities of LLMs is to report their performance, e.g., accuracy, on end tasks that require reasoning.
We list some common benchmarks as follows.

\p{Arithmetic Reasoning}
\textit{Arithmetic reasoning} is the ability to understand and apply mathematical concepts and principles in order to solve problems involving arithmetic operations. 
This involves using logical thinking and mathematical principles to determine the correct course of action when solving mathematical problems. 
Representative benchmarks for arithmetic reasoning include GSM8K~\citep{cobbe2021training},
Math~\citep{dan2021@measuring},
MathQA~\citep{amini2019mathqa}, 
SVAMP~\citep{patel2021nlp},
ASDiv~\citep{miao2020diverse},
AQuA~\citep{ling2017program},
and MAWPS~\citep{roy2015solving}. 
It is worth mentioning that
\citet{anil2022exploring} generate the \textit{Parity Datasets} and the \textit{Boolean Variable Assignment Dataset} for analyzing the length generalization capabilities of LLMs (\S\ref{sec:trainig_and_prompting}).

\p{Commonsense Reasoning}
\textit{Commonsense Reasoning} is the use of everyday knowledge and understanding to make judgments and predictions about new situations. It is a fundamental aspect of human intelligence that enables us to navigate our environment, understand others, and make decisions with incomplete information.
Benchmarks that can be used for testing commonsense reasoning abilities of LLMs include
CSQA~\citep{talmor2019commonsenseqa}, StrategyQA~\citep{geva2021did}, and ARC~\citep{clark2018think}.
We refer the reader to \citet{bhargava2022commonsense}'s survey for more work in this domain.

\p{Symbolic Reasoning}
\textit{Symbolic reasoning} is a form of reasoning that involves the manipulation of symbols according to formal rules.
In symbolic reasoning, we use abstract symbols to represent concepts and relationships, and then manipulate those symbols according to precise rules in order to draw conclusions or solve problems. 
Two benchmarks of symbolic reasoning are presented in \citet{wei2022chain}, including 
Last Letter Concatenation and Coin Flip.

\p{Others}
In practice, there are many benchmarks that can be used to evaluate reasoning abilities of LLMs (indirectly), as long as the downstream task involves reasoning.
BIG-bench~\citep{srivastava2022beyond}, for example, includes over 200 tasks that test a range of reasoning skills, including tasks like Date Understanding, Word Sorting, and Causal Judgement.
Other benchmarks, such as SCAN~\citep{lake2018generalization} and the one proposed by \citet{anil2022exploring}, focus on evaluating generalization ability. 
LLMs can also be tested on their table reasoning abilities using benchmarks such as WikiTableQA~\citep{pasupat2015compositional}, FetaQA~\citep{nan2022fetaqa}, as suggested by \citet{chen2022large}.
In addition, there are benchmarks for evaluating LLMs' generative relational reasoning abilities, such as 
CommonGen~\citep{lin2020commongen,liu2022dimongen} and Open Relation Modeling~\citep{huang2022open,huang2022deer}.

\subsection{Analysis on Reasoning}
\label{sec:analysis}

Although LLMs have demonstrated impressive performance on various reasoning tasks, the extent to which their predictions are based on true reasoning or simple heuristics is not always clear.
This is because most existing evaluations focus on their accuracy on end tasks, rather than directly assessing their reasoning steps.
While some error analysis has been conducted on the generated rationales of LLMs \citep[][\textit{inter alia}]{wei2022chain,kojima2022large}, this analysis has often been limited in depth.

There have been some efforts to develop metrics and benchmarks that enable a more formal/deep analysis of reasoning in LLMs.
\citet{golovneva2022roscoe} design ROSCOE, a set of interpretable, detailed step-by-step evaluation metrics covering various perspectives including semantic alignment, logical inference, semantic similarity, and language coherence.
\citet{saparov2022language} create a synthetic dataset called PrOntoQA that is generated from real or fictional ontologies. Each example in the dataset has a unique proof, which can be converted to simple sentences and back again, allowing for a formal analysis of each reasoning step. 
\citet{han2022folio} introduce a dataset called FOLIO to test the first-order logic reasoning capabilities of LLMs. FOLIO contains first-order logic reasoning problems that require models to determine the correctness of conclusions given a set of premises.
In addition, \citet{wang2022towards} conduct ablation experiments on CoT and find that LLMs may also perform reasoning while prompting with invalid rationals. Their study also suggests that being relevant to the query and correctly ordering the reasoning steps are important for CoT prompting.

In summary, most existing studies primarily report the performance of the models on downstream reasoning tasks, without a detailed examination of the quality of the rationales produced. This leaves open the question of whether the models are actually able to reason in a way that is similar to human reasoning, or whether they are simply able to achieve good performance on the tasks through other means. Further research is needed to more formally analyze the reasoning abilities of LLMs.

\section{Findings and Implications}
\label{sec:findings}

In this section, we summarize the important findings and implications of studies on reasoning in large language models.

\p{Reasoning seems an emergent ability of LLMs}
\citet{wei2022emergent,wei2022chain,suzgun2022challenging} show that 
reasoning ability appears to emerge only in large language models like GPT-3 175B, as evidenced by significant improvements in performance on reasoning tasks at a certain scale (e.g., 100 billion parameters).
This suggests that it may be more effective to utilize large models for general reasoning problems rather than training small models for specific tasks. 
However, the reason for this emergent ability is not yet fully understood. We refer the reader to \citet{wei2022emergent,fu_khot_peng_2022b} for some potential explanations.

\p{Chain of thought elicits ``reasoning'' of LLMs}
The use of chain-of-thought (CoT) prompts~\citep{wei2022chain} has been shown to improve the performance of LLMs on various reasoning tasks, as demonstrated in the experiments of \citet{wei2022emergent,wei2022chain,suzgun2022challenging}.
Additionally, \citet{saparov2022language} (\S\ref{sec:analysis}) find that, when using CoT prompts, LLMs are able to produce valid individual proof steps, even when the synthetic ontology is fictional or counterfactual. However, they may sometimes choose the wrong steps when multiple options are available, leading to incomplete or incorrect proofs.
Moreover, for many reasoning tasks where the performance of standard prompting grows smoothly with model scale, chain-of-thought prompting can lead to dramatic performance improvement.
In addition to these benefits, the use of CoT prompts has been shown to improve the out-of-distribution robustness of LLMs \citep[][\textit{inter alia}]{wei2022chain,zhou2022least,anil2022exploring}, an advantage that is not typically observed with standard prompting or fully supervised finetuning paradigms.

\p{LLMs show human-like content effects on reasoning}
According to \citet{dasgupta2022language}, LLMs exhibit reasoning patterns that are similar to those of humans as described in the cognitive literature.
For example, the models' predictions are influenced by both prior knowledge and abstract reasoning, and their judgments of logical validity are impacted by the believability of the conclusions. These findings suggest that, although language models may not always perform well on reasoning tasks, their failures often occur in situations that are challenging for humans as well. This provides some evidence that language models may ``reason'' in a way that is similar to human reasoning.

\p{LLMs are still unskilled at complex reasoning}
Although LLMs seem to possess impressive reasoning capabilities with the techniques described in \S\ref{sec:method}, they still struggle with more complex reasoning tasks or those involving implicature, according to studies such as \citet{valmeekam2022large,han2022folio,ruis2022large}.
For instance, \citet{valmeekam2022large} find that even in relatively simple commonsense planning domains that humans would have no trouble navigating, LLMs such as GPT-3~\citep{brown2020language} and BLOOM~\citep{scao2022bloom} struggle to perform effectively.
These findings suggest that existing benchmarks may be too simple to accurately gauge the true reasoning abilities of LLMs, and that more challenging tasks may be needed to fully evaluate their abilities in this regard.

\section{Reflection, Discussion, and Future Directions}
\label{sec:discussion}

\ps{Why reasoning}
Reasoning is the process of thinking about something in a logical and systematic way, and it is a key aspect of human intelligence. By incorporating reasoning capabilities into language models, we can enable them to perform tasks that require more complex and nuanced thinking, such as problem solving, decision making, and planning~\citep{pmlr-v162-huang22a,huanginner,song2022llm}.
This can improve the performance of these models on downstream tasks and increase their out-of-distribution robustness \cite{wei2022emergent,wei2022chain,suzgun2022challenging,zhou2022least,anil2022exploring}. In addition, reasoning can make language models more explainable and interpretable, as it provides explicit rationales for their predictions.

\ps{Right task/application} As \citet{valmeekam2022large} point out, current benchmarks may not adequately reflect the reasoning capabilities of LLMs.
In addition, tasks such as solving simple math problems and concatenating letters in strings (\S\ref{sec:endtask}) are artificial and do not accurately reflect real-world situations. To truly understand the reasoning ability of LLMs, it is important to consider more realistic and meaningful applications such as decision making~\citep{edwards1954theory}, legal reasoning~\citep{levi2013introduction}, and scientific reasoning~\citep{zimmerman2000development}.
Our ultimate goal should not be to enable LLMs to solve simple math problems, which can be simply done with other programs.
When conducting relevant research, it is essential to ask
\textit{whether the specific task being tackled is meaningful} and \textit{whether the proposed method can be generalized to more realistic tasks and applications}.

\ps{Are language models really able to reason}
There are several indications that LLMs are able to reason, including 1) high performance on various tasks requiring reasoning \citep{suzgun2022challenging}; 2) the ability to reason step-by-step with chain-of-thought prompting \citep{wei2022chain}; and 3) the reflection of human-like content effects on reasoning \citep{dasgupta2022language}. 
However, these findings are not sufficient to conclude that LLMs can truly reason. For 1), it is not clear whether the models are making predictions based on \textit{reasoning} or \textit{heuristics} \citep{patel2021nlp}. For many existing benchmarks on reasoning, actually, we can design a program with heuristic rules to achieve very high performance. 
We usually do not think a program relying on heuristic rules is capable of reasoning.
For 2), although the models seem to reason step-by-step, the generated rationales may be incorrect and inconsistent. It is possible that the models are ``generating reasoning-like response'' rather than ``reasoning step-by-step''. For 3), while LLMs display some human-like reasoning patterns, this does not necessarily mean that they behave like humans.

Additionally, there are several observations that suggest LLMs may not be capable of reasoning: 1) LLMs still struggle with tasks that require complex reasoning \citep{valmeekam2022large,han2022folio,ruis2022large}. If LLMs are really decent reasoners, they should handle tasks that can be simply solved by humans through reasoning;
2) LLMs make mistakes in their reasoning, as explained above;
3)$^\#$\footnote{~$^\#$indicates the finding has not been carefully examined in language models with more than 100 billion parameters.} The performance of LLMs on downstream tasks has been found to be sensitive to the frequency of certain terms, such as numbers, in the training data  \citep{razeghi2022impact,jung2022maieutic}, which would not be expected if the models were solving mathematical problems through reasoning; 4)$^\#$ Language models have been found to struggle with associating relevant information that they have memorized \citep{huang2022large}.

Overall, it is still too early to draw a conclusion about the proposed question. 
In fact, there is also an ongoing debate about whether language models can actually \textit{understand} language or capture \textit{meaning} \citep{bender2020climbing,li2021implicit,manning2022human,piantasodi2022meaning}. 
Further in-depth analysis of factors such as training data, model architecture, and optimization objectives is needed, as well as the development of better benchmarks for measuring the reasoning capabilities of LLMs. However, it is clear that the current models are not yet capable of robust reasoning.

\p{Improving reasoning capabilities of LLMs}
While techniques like chain-of-thought prompting~\citep{wei2022chain} may help to elicit reasoning abilities in large language models, they cannot enable the models to solve tasks beyond their current capabilities.
To truly enhance reasoning in LLMs, we need to utilize training data, model architecture, and optimization objectives that are designed to encourage reasoning.
For example, finetuning a model with a dataset including CoT data has been shown to improve reasoning \citep{chung2022scaling}, and models can also self-improve through the process of bootstrapping their reasoning \citep{zelikman2022star,huang2022large2}.
There is still much research that needs to be done in this area, and we look forward to future progress in improving reasoning in large language models.

\section{Conclusion}
In this paper, we have provided a detailed and up-to-date review of the current state of knowledge on reasoning in large language models. We have discussed techniques for improving and eliciting reasoning in LLMs, methods and benchmarks for evaluating reasoning abilities, and the findings and implications of previous studies in this topic. 
While LLMs have made significant progress in natural language processing and related fields, it remains unclear to what extent they are capable of true reasoning or whether they are simply using memorized patterns and heuristics to solve problems. Further research is needed to fully understand the reasoning abilities of LLMs, improve LLMs' reasoning capabilities, and determine their potential for use in a variety of applications. 
We hope that this paper will serve as a useful overview of the current state of the field and stimulate further discussion and research on this interesting and important topic.

\section*{Limitations}

In this paper, we provide an overview of the current state of knowledge on reasoning in large language models. Reasoning is a broad concept that encompasses various forms, making it impractical to summarize all related work in a single paper. Therefore, we focus on deductive reasoning, as it is the most commonly studied in the literature. Other forms of reasoning such as inductive reasoning~\citep[][\textit{inter alia}]{yang2022language,misra2022property} and abductive reasoning~\citep[][\textit{inter alia}]{wiegreffe2021reframing,lampinen2022can,jung2022maieutic} may not be discussed in depth.

Additionally, given the rapid evolution and significance of reasoning within large language models, it is crucial to note that new contributions may have emerged in the field concurrent with the writing of this paper. An additional resource to consider is a parallel survey by \citet{qiao2022reasoning}, which emphasizes reasoning via language model prompting. Our coverage may not extend to papers released during or after 2023 such as evaluation on ChatGPT~\citep{bang2023multitask,zheng2023does}. As such, we recommend readers to check the papers that cite this survey for a more comprehensive and updated understanding of this field.

\section*{Acknowledgements}

We would like to thank Jason Wei (OpenAI) and Denny Zhou (Google DeepMind) for their valuable advice and constructive feedback on this work.
This material is based upon work supported by the National Science Foundation IIS 16-19302 and IIS 16-33755, Zhejiang University ZJU Research 083650, IBM-Illinois Center for Cognitive Computing Systems Research (C3SR) and IBM-Illinois Discovery Accelerator Institute (IIDAI), gift grants from eBay and Microsoft Azure, UIUC OVCR CCIL Planning Grant 434S34, UIUC CSBS Small Grant 434C8U, and UIUC New Frontiers Initiative. Any opinions, findings, and conclusions or recommendations expressed in this publication are those of the author(s) and do not necessarily reflect the views of the funding agencies.

\bibliography{custom}

\begin{thebibliography}{122}
\expandafter\ifx\csname natexlab\endcsname\relax\def\natexlab#1{#1}\fi

\bibitem[{Amini et~al.(2019)Amini, Gabriel, Lin, Koncel-Kedziorski, Choi, and
  Hajishirzi}]{amini2019mathqa}
Aida Amini, Saadia Gabriel, Shanchuan Lin, Rik Koncel-Kedziorski, Yejin Choi,
  and Hannaneh Hajishirzi. 2019.
\newblock \href {https://doi.org/10.18653/v1/N19-1245} {{M}ath{QA}: Towards
  interpretable math word problem solving with operation-based formalisms}.
\newblock In \emph{Proceedings of the 2019 Conference of the North {A}merican
  Chapter of the Association for Computational Linguistics: Human Language
  Technologies, Volume 1 (Long and Short Papers)}, pages 2357--2367,
  Minneapolis, Minnesota. Association for Computational Linguistics.

\bibitem[{Anil et~al.(2022)Anil, Wu, Andreassen, Lewkowycz, Misra, Ramasesh,
  Slone, Gur-Ari, Dyer, and Neyshabur}]{anil2022exploring}
Cem Anil, Yuhuai Wu, Anders Andreassen, Aitor Lewkowycz, Vedant Misra, Vinay
  Ramasesh, Ambrose Slone, Guy Gur-Ari, Ethan Dyer, and Behnam Neyshabur. 2022.
\newblock \href {https://arxiv.org/abs/2207.04901} {Exploring length
  generalization in large language models}.
\newblock \emph{ArXiv preprint}, abs/2207.04901.

\bibitem[{Bang et~al.(2023)Bang, Cahyawijaya, Lee, Dai, Su, Wilie, Lovenia, Ji,
  Yu, Chung et~al.}]{bang2023multitask}
Yejin Bang, Samuel Cahyawijaya, Nayeon Lee, Wenliang Dai, Dan Su, Bryan Wilie,
  Holy Lovenia, Ziwei Ji, Tiezheng Yu, Willy Chung, et~al. 2023.
\newblock \href {https://arxiv.org/abs/2302.04023} {A multitask, multilingual,
  multimodal evaluation of chatgpt on reasoning, hallucination, and
  interactivity}.
\newblock \emph{ArXiv preprint}, abs/2302.04023.

\bibitem[{Bender and Koller(2020)}]{bender2020climbing}
Emily~M. Bender and Alexander Koller. 2020.
\newblock \href {https://doi.org/10.18653/v1/2020.acl-main.463} {Climbing
  towards {NLU}: {On} meaning, form, and understanding in the age of data}.
\newblock In \emph{Proceedings of the 58th Annual Meeting of the Association
  for Computational Linguistics}, pages 5185--5198, Online. Association for
  Computational Linguistics.

\bibitem[{Bhargava and Ng(2022)}]{bhargava2022commonsense}
Prajjwal Bhargava and Vincent Ng. 2022.
\newblock Commonsense knowledge reasoning and generation with pre-trained
  language models: A survey.
\newblock \emph{Proceedings of the AAAI Conference on Artificial Intelligence}.

\bibitem[{Bommasani et~al.(2021)Bommasani, Hudson, Adeli, Altman, Arora, von
  Arx, Bernstein, Bohg, Bosselut, Brunskill
  et~al.}]{bommasani2021opportunities}
Rishi Bommasani, Drew~A Hudson, Ehsan Adeli, Russ Altman, Simran Arora, Sydney
  von Arx, Michael~S Bernstein, Jeannette Bohg, Antoine Bosselut, Emma
  Brunskill, et~al. 2021.
\newblock \href {https://arxiv.org/abs/2108.07258} {On the opportunities and
  risks of foundation models}.
\newblock \emph{ArXiv preprint}, abs/2108.07258.

\bibitem[{Bronkhorst et~al.(2020)Bronkhorst, Roorda, Suhre, and
  Goedhart}]{bronkhorst2020logical}
Hugo Bronkhorst, Gerrit Roorda, Cor Suhre, and Martin Goedhart. 2020.
\newblock Logical reasoning in formal and everyday reasoning tasks.
\newblock \emph{International Journal of Science and Mathematics Education},
  18(8):1673--1694.

\bibitem[{Brown et~al.(2020)Brown, Mann, Ryder, Subbiah, Kaplan, Dhariwal,
  Neelakantan, Shyam, Sastry, Askell, Agarwal, Herbert{-}Voss, Krueger,
  Henighan, Child, Ramesh, Ziegler, Wu, Winter, Hesse, Chen, Sigler, Litwin,
  Gray, Chess, Clark, Berner, McCandlish, Radford, Sutskever, and
  Amodei}]{brown2020language}
Tom~B. Brown, Benjamin Mann, Nick Ryder, Melanie Subbiah, Jared Kaplan,
  Prafulla Dhariwal, Arvind Neelakantan, Pranav Shyam, Girish Sastry, Amanda
  Askell, Sandhini Agarwal, Ariel Herbert{-}Voss, Gretchen Krueger, Tom
  Henighan, Rewon Child, Aditya Ramesh, Daniel~M. Ziegler, Jeffrey Wu, Clemens
  Winter, Christopher Hesse, Mark Chen, Eric Sigler, Mateusz Litwin, Scott
  Gray, Benjamin Chess, Jack Clark, Christopher Berner, Sam McCandlish, Alec
  Radford, Ilya Sutskever, and Dario Amodei. 2020.
\newblock \href
  {https://proceedings.neurips.cc/paper/2020/hash/1457c0d6bfcb4967418bfb8ac142f64a-Abstract.html}
  {Language models are few-shot learners}.
\newblock In \emph{Advances in Neural Information Processing Systems 33: Annual
  Conference on Neural Information Processing Systems 2020, NeurIPS 2020,
  December 6-12, 2020, virtual}.

\bibitem[{Chen et~al.(2021)Chen, Tworek, Jun, Yuan, Pinto, Kaplan, Edwards,
  Burda, Joseph, Brockman et~al.}]{chen2021evaluating}
Mark Chen, Jerry Tworek, Heewoo Jun, Qiming Yuan, Henrique Ponde de~Oliveira
  Pinto, Jared Kaplan, Harri Edwards, Yuri Burda, Nicholas Joseph, Greg
  Brockman, et~al. 2021.
\newblock \href {https://arxiv.org/abs/2107.03374} {Evaluating large language
  models trained on code}.
\newblock \emph{ArXiv preprint}, abs/2107.03374.

\bibitem[{Chen(2022)}]{chen2022large}
Wenhu Chen. 2022.
\newblock \href {https://arxiv.org/abs/2210.06710} {Large language models are
  few (1)-shot table reasoners}.
\newblock \emph{ArXiv preprint}, abs/2210.06710.

\bibitem[{Chen et~al.(2022)Chen, Ma, Wang, and Cohen}]{chen2022program}
Wenhu Chen, Xueguang Ma, Xinyi Wang, and William~W Cohen. 2022.
\newblock \href {https://arxiv.org/abs/2211.12588} {Program of thoughts
  prompting: Disentangling computation from reasoning for numerical reasoning
  tasks}.
\newblock \emph{ArXiv preprint}, abs/2211.12588.

\bibitem[{Chowdhery et~al.(2022)Chowdhery, Narang, Devlin, Bosma, Mishra,
  Roberts, Barham, Chung, Sutton, Gehrmann et~al.}]{chowdhery2022palm}
Aakanksha Chowdhery, Sharan Narang, Jacob Devlin, Maarten Bosma, Gaurav Mishra,
  Adam Roberts, Paul Barham, Hyung~Won Chung, Charles Sutton, Sebastian
  Gehrmann, et~al. 2022.
\newblock \href {https://arxiv.org/abs/2204.02311} {Palm: Scaling language
  modeling with pathways}.
\newblock \emph{ArXiv preprint}, abs/2204.02311.

\bibitem[{Chung et~al.(2022)Chung, Hou, Longpre, Zoph, Tay, Fedus, Li, Wang,
  Dehghani, Brahma et~al.}]{chung2022scaling}
Hyung~Won Chung, Le~Hou, Shayne Longpre, Barret Zoph, Yi~Tay, William Fedus,
  Eric Li, Xuezhi Wang, Mostafa Dehghani, Siddhartha Brahma, et~al. 2022.
\newblock \href {https://arxiv.org/abs/2210.11416} {Scaling
  instruction-finetuned language models}.
\newblock \emph{ArXiv preprint}, abs/2210.11416.

\bibitem[{Clark et~al.(2018)Clark, Cowhey, Etzioni, Khot, Sabharwal, Schoenick,
  and Tafjord}]{clark2018think}
Peter Clark, Isaac Cowhey, Oren Etzioni, Tushar Khot, Ashish Sabharwal, Carissa
  Schoenick, and Oyvind Tafjord. 2018.
\newblock \href {https://arxiv.org/abs/1803.05457} {Think you have solved
  question answering? try arc, the ai2 reasoning challenge}.
\newblock \emph{ArXiv preprint}, abs/1803.05457.

\bibitem[{Cobbe et~al.(2021)Cobbe, Kosaraju, Bavarian, Hilton, Nakano, Hesse,
  and Schulman}]{cobbe2021training}
Karl Cobbe, Vineet Kosaraju, Mohammad Bavarian, Jacob Hilton, Reiichiro Nakano,
  Christopher Hesse, and John Schulman. 2021.
\newblock \href {https://arxiv.org/abs/2110.14168} {Training verifiers to solve
  math word problems}.
\newblock \emph{ArXiv preprint}, abs/2110.14168.

\bibitem[{Creswell and Shanahan(2022)}]{creswell2022faithful}
Antonia Creswell and Murray Shanahan. 2022.
\newblock \href {https://arxiv.org/abs/2208.14271} {Faithful reasoning using
  large language models}.
\newblock \emph{ArXiv preprint}, abs/2208.14271.

\bibitem[{Creswell et~al.(2022)Creswell, Shanahan, and
  Higgins}]{creswell2022selection}
Antonia Creswell, Murray Shanahan, and Irina Higgins. 2022.
\newblock \href {https://arxiv.org/abs/2205.09712} {Selection-inference:
  Exploiting large language models for interpretable logical reasoning}.
\newblock \emph{ArXiv preprint}, abs/2205.09712.

\bibitem[{Dasgupta et~al.(2022)Dasgupta, Lampinen, Chan, Creswell, Kumaran,
  McClelland, and Hill}]{dasgupta2022language}
Ishita Dasgupta, Andrew~K Lampinen, Stephanie~CY Chan, Antonia Creswell,
  Dharshan Kumaran, James~L McClelland, and Felix Hill. 2022.
\newblock \href {https://arxiv.org/abs/2207.07051} {Language models show
  human-like content effects on reasoning}.
\newblock \emph{ArXiv preprint}, abs/2207.07051.

\bibitem[{Devlin et~al.(2019)Devlin, Chang, Lee, and
  Toutanova}]{kenton2019bert}
Jacob Devlin, Ming-Wei Chang, Kenton Lee, and Kristina Toutanova. 2019.
\newblock \href {https://doi.org/10.18653/v1/N19-1423} {{BERT}: Pre-training of
  deep bidirectional transformers for language understanding}.
\newblock In \emph{Proceedings of the 2019 Conference of the North {A}merican
  Chapter of the Association for Computational Linguistics: Human Language
  Technologies, Volume 1 (Long and Short Papers)}, pages 4171--4186,
  Minneapolis, Minnesota. Association for Computational Linguistics.

\bibitem[{Dohan et~al.(2022)Dohan, Xu, Lewkowycz, Austin, Bieber, Lopes, Wu,
  Michalewski, Saurous, Sohl-Dickstein et~al.}]{dohan2022language}
David Dohan, Winnie Xu, Aitor Lewkowycz, Jacob Austin, David Bieber,
  Raphael~Gontijo Lopes, Yuhuai Wu, Henryk Michalewski, Rif~A Saurous, Jascha
  Sohl-Dickstein, et~al. 2022.
\newblock \href {https://arxiv.org/abs/2207.10342} {Language model cascades}.
\newblock \emph{ArXiv preprint}, abs/2207.10342.

\bibitem[{Drozdov et~al.(2022)Drozdov, Sch{\"a}rli, Aky{\"u}rek, Scales, Song,
  Chen, Bousquet, and Zhou}]{drozdov2022compositional}
Andrew Drozdov, Nathanael Sch{\"a}rli, Ekin Aky{\"u}rek, Nathan Scales, Xinying
  Song, Xinyun Chen, Olivier Bousquet, and Denny Zhou. 2022.
\newblock \href {https://arxiv.org/abs/2209.15003} {Compositional semantic
  parsing with large language models}.
\newblock \emph{ArXiv preprint}, abs/2209.15003.

\bibitem[{Dua et~al.(2022)Dua, Gupta, Singh, and Gardner}]{dua2022successive}
Dheeru Dua, Shivanshu Gupta, Sameer Singh, and Matt Gardner. 2022.
\newblock \href {https://arxiv.org/abs/2212.04092} {Successive prompting for
  decomposing complex questions}.
\newblock \emph{ArXiv preprint}, abs/2212.04092.

\bibitem[{Edwards(1954)}]{edwards1954theory}
Ward Edwards. 1954.
\newblock The theory of decision making.
\newblock \emph{Psychological bulletin}, 51(4):380.

\bibitem[{Fagin et~al.(2004)Fagin, Halpern, Moses, and
  Vardi}]{fagin2004reasoning}
Ronald Fagin, Joseph~Y Halpern, Yoram Moses, and Moshe Vardi. 2004.
\newblock \emph{Reasoning about knowledge}.
\newblock MIT press.

\bibitem[{Fu et~al.(2022{\natexlab{a}})Fu, Peng, and Khot}]{fu_khot_peng_2022b}
Yao Fu, Hao Peng, and Tushar Khot. 2022{\natexlab{a}}.
\newblock \href
  {https://yaofu.notion.site/How-does-GPT-Obtain-its-Ability-Tracing-Emergent-Abilities-of-Language-Models-to-their-Sources-b9a57ac0fcf74f30a1ab9e3e36fa1dc1}
  {How does gpt obtain its ability? tracing emergent abilities of language
  models to their sources}.

\bibitem[{Fu et~al.(2022{\natexlab{b}})Fu, Peng, Sabharwal, Clark, and
  Khot}]{fu2022complexity}
Yao Fu, Hao Peng, Ashish Sabharwal, Peter Clark, and Tushar Khot.
  2022{\natexlab{b}}.
\newblock \href {https://arxiv.org/abs/2210.00720} {Complexity-based prompting
  for multi-step reasoning}.
\newblock \emph{ArXiv preprint}, abs/2210.00720.

\bibitem[{Galotti(1989)}]{galotti1989approaches}
Kathleen~M Galotti. 1989.
\newblock Approaches to studying formal and everyday reasoning.
\newblock \emph{Psychological bulletin}, 105(3):331.

\bibitem[{Gao et~al.(2022)Gao, Madaan, Zhou, Alon, Liu, Yang, Callan, and
  Neubig}]{gao2022pal}
Luyu Gao, Aman Madaan, Shuyan Zhou, Uri Alon, Pengfei Liu, Yiming Yang, Jamie
  Callan, and Graham Neubig. 2022.
\newblock \href {https://arxiv.org/abs/2211.10435} {Pal: Program-aided language
  models}.
\newblock \emph{ArXiv preprint}, abs/2211.10435.

\bibitem[{Geva et~al.(2021)Geva, Khashabi, Segal, Khot, Roth, and
  Berant}]{geva2021did}
Mor Geva, Daniel Khashabi, Elad Segal, Tushar Khot, Dan Roth, and Jonathan
  Berant. 2021.
\newblock \href {https://doi.org/10.1162/tacl_a_00370} {Did aristotle use a
  laptop? a question answering benchmark with implicit reasoning strategies}.
\newblock \emph{Transactions of the Association for Computational Linguistics},
  9:346--361.

\bibitem[{Golovneva et~al.(2022)Golovneva, Chen, Poff, Corredor, Zettlemoyer,
  Fazel-Zarandi, and Celikyilmaz}]{golovneva2022roscoe}
Olga Golovneva, Moya Chen, Spencer Poff, Martin Corredor, Luke Zettlemoyer,
  Maryam Fazel-Zarandi, and Asli Celikyilmaz. 2022.
\newblock \href {https://arxiv.org/abs/2212.07919} {Roscoe: A suite of metrics
  for scoring step-by-step reasoning}.
\newblock \emph{ArXiv preprint}, abs/2212.07919.

\bibitem[{Han et~al.(2022{\natexlab{a}})Han, Schoelkopf, Zhao, Qi, Riddell,
  Benson, Sun, Zubova, Qiao, Burtell et~al.}]{han2022folio}
Simeng Han, Hailey Schoelkopf, Yilun Zhao, Zhenting Qi, Martin Riddell, Luke
  Benson, Lucy Sun, Ekaterina Zubova, Yujie Qiao, Matthew Burtell, et~al.
  2022{\natexlab{a}}.
\newblock \href {https://arxiv.org/abs/2209.00840} {Folio: Natural language
  reasoning with first-order logic}.
\newblock \emph{ArXiv preprint}, abs/2209.00840.

\bibitem[{Han et~al.(2022{\natexlab{b}})Han, Ransom, Perfors, and
  Kemp}]{han2022human}
Simon~Jerome Han, Keith Ransom, Andrew Perfors, and Charles Kemp.
  2022{\natexlab{b}}.
\newblock Human-like property induction is a challenge for large language
  models.

\bibitem[{He et~al.(2023)He, Zhang, and Roth}]{he2022rethinking}
Hangfeng He, Hongming Zhang, and Dan Roth. 2023.
\newblock \href {https://arxiv.org/abs/2301.00303} {Rethinking with retrieval:
  Faithful large language model inference}.
\newblock \emph{ArXiv preprint}, abs/2301.00303.

\bibitem[{Helwe et~al.(2021)Helwe, Clavel, and Suchanek}]{helwe2021reasoning}
Chadi Helwe, Chlo{\'e} Clavel, and Fabian~M Suchanek. 2021.
\newblock Reasoning with transformer-based models: Deep learning, but shallow
  reasoning.
\newblock In \emph{3rd Conference on Automated Knowledge Base Construction}.

\bibitem[{Hendrycks et~al.(2021)Hendrycks, Burns, Kadavath, Arora, Basart,
  Tang, Song, and Steinhardt}]{dan2021@measuring}
Dan Hendrycks, Collin Burns, Saurav Kadavath, Akul Arora, Steven Basart, Eric
  Tang, Dawn Song, and Jacob Steinhardt. 2021.
\newblock \href
  {https://datasets-benchmarks-proceedings.neurips.cc/paper/2021/file/be83ab3ecd0db773eb2dc1b0a17836a1-Paper-round2.pdf}
  {Measuring mathematical problem solving with the math dataset}.
\newblock In \emph{Proceedings of the Neural Information Processing Systems
  Track on Datasets and Benchmarks}, volume~1.

\bibitem[{Huang et~al.(2022{\natexlab{a}})Huang, Gu, Hou, Wu, Wang, Yu, and
  Han}]{huang2022large2}
Jiaxin Huang, Shixiang~Shane Gu, Le~Hou, Yuexin Wu, Xuezhi Wang, Hongkun Yu,
  and Jiawei Han. 2022{\natexlab{a}}.
\newblock \href {https://arxiv.org/abs/2210.11610} {Large language models can
  self-improve}.
\newblock \emph{ArXiv preprint}, abs/2210.11610.

\bibitem[{Huang et~al.(2022{\natexlab{b}})Huang, Chang, Xiong, and
  Hwu}]{huang2022open}
Jie Huang, Kevin Chang, Jinjun Xiong, and Wen-mei Hwu. 2022{\natexlab{b}}.
\newblock \href {https://doi.org/10.18653/v1/2022.findings-acl.26} {Open
  relation modeling: Learning to define relations between entities}.
\newblock In \emph{Findings of the Association for Computational Linguistics:
  ACL 2022}, pages 297--308, Dublin, Ireland. Association for Computational
  Linguistics.

\bibitem[{Huang et~al.(2022{\natexlab{c}})Huang, Shao, and
  Chang}]{huang2022large}
Jie Huang, Hanyin Shao, and Kevin Chen-Chuan Chang. 2022{\natexlab{c}}.
\newblock \href {https://aclanthology.org/2022.findings-emnlp.148} {Are large
  pre-trained language models leaking your personal information?}
\newblock In \emph{Findings of the Association for Computational Linguistics:
  EMNLP 2022}, pages 2038--2047, Abu Dhabi, United Arab Emirates. Association
  for Computational Linguistics.

\bibitem[{Huang et~al.(2022{\natexlab{d}})Huang, Zhu, Chang, Xiong, and
  Hwu}]{huang2022deer}
Jie Huang, Kerui Zhu, Kevin Chen-Chuan Chang, Jinjun Xiong, and Wen-mei Hwu.
  2022{\natexlab{d}}.
\newblock \href {https://aclanthology.org/2022.emnlp-main.448} {{DEER}:
  Descriptive knowledge graph for explaining entity relationships}.
\newblock In \emph{Proceedings of the 2022 Conference on Empirical Methods in
  Natural Language Processing}, pages 6686--6698, Abu Dhabi, United Arab
  Emirates. Association for Computational Linguistics.

\bibitem[{Huang et~al.(2022{\natexlab{e}})Huang, Abbeel, Pathak, and
  Mordatch}]{pmlr-v162-huang22a}
Wenlong Huang, Pieter Abbeel, Deepak Pathak, and Igor Mordatch.
  2022{\natexlab{e}}.
\newblock \href {https://proceedings.mlr.press/v162/huang22a.html} {Language
  models as zero-shot planners: Extracting actionable knowledge for embodied
  agents}.
\newblock In \emph{Proceedings of the 39th International Conference on Machine
  Learning}, volume 162 of \emph{Proceedings of Machine Learning Research},
  pages 9118--9147. PMLR.

\bibitem[{Huang et~al.(2022{\natexlab{f}})Huang, Xia, Xiao, Chan, Liang,
  Florence, Zeng, Tompson, Mordatch, Chebotar et~al.}]{huanginner}
Wenlong Huang, Fei Xia, Ted Xiao, Harris Chan, Jacky Liang, Pete Florence, Andy
  Zeng, Jonathan Tompson, Igor Mordatch, Yevgen Chebotar, et~al.
  2022{\natexlab{f}}.
\newblock Inner monologue: Embodied reasoning through planning with language
  models.
\newblock In \emph{2022 Conference on Robot Learning}.

\bibitem[{Huth and Ryan(2004)}]{huth2004logic}
Michael Huth and Mark Ryan. 2004.
\newblock \emph{Logic in Computer Science: Modelling and reasoning about
  systems}.
\newblock Cambridge university press.

\bibitem[{Jung et~al.(2022)Jung, Qin, Welleck, Brahman, Bhagavatula, Bras, and
  Choi}]{jung2022maieutic}
Jaehun Jung, Lianhui Qin, Sean Welleck, Faeze Brahman, Chandra Bhagavatula,
  Ronan~Le Bras, and Yejin Choi. 2022.
\newblock Maieutic prompting: Logically consistent reasoning with recursive
  explanations.
\newblock \emph{The 2022 Conference on Empirical Methods for Natural Language
  Processing}.

\bibitem[{Kazemi et~al.(2022)Kazemi, Kim, Bhatia, Xu, and
  Ramachandran}]{kazemi2022lambada}
Seyed~Mehran Kazemi, Najoung Kim, Deepti Bhatia, Xin Xu, and Deepak
  Ramachandran. 2022.
\newblock \href {https://arxiv.org/abs/2212.13894} {Lambada: Backward chaining
  for automated reasoning in natural language}.
\newblock \emph{ArXiv preprint}, abs/2212.13894.

\bibitem[{Keysers et~al.(2020)Keysers, Sch{\"{a}}rli, Scales, Buisman, Furrer,
  Kashubin, Momchev, Sinopalnikov, Stafiniak, Tihon, Tsarkov, Wang, van Zee,
  and Bousquet}]{keysers2019measuring}
Daniel Keysers, Nathanael Sch{\"{a}}rli, Nathan Scales, Hylke Buisman, Daniel
  Furrer, Sergii Kashubin, Nikola Momchev, Danila Sinopalnikov, Lukasz
  Stafiniak, Tibor Tihon, Dmitry Tsarkov, Xiao Wang, Marc van Zee, and Olivier
  Bousquet. 2020.
\newblock \href {https://openreview.net/forum?id=SygcCnNKwr} {Measuring
  compositional generalization: {A} comprehensive method on realistic data}.
\newblock In \emph{8th International Conference on Learning Representations,
  {ICLR} 2020, Addis Ababa, Ethiopia, April 26-30, 2020}. OpenReview.net.

\bibitem[{Khot et~al.(2022)Khot, Trivedi, Finlayson, Fu, Richardson, Clark, and
  Sabharwal}]{khot2022decomposed}
Tushar Khot, Harsh Trivedi, Matthew Finlayson, Yao Fu, Kyle Richardson, Peter
  Clark, and Ashish Sabharwal. 2022.
\newblock \href {https://arxiv.org/abs/2210.02406} {Decomposed prompting: A
  modular approach for solving complex tasks}.
\newblock \emph{ArXiv preprint}, abs/2210.02406.

\bibitem[{Kojima et~al.(2022)Kojima, Gu, Reid, Matsuo, and
  Iwasawa}]{kojima2022large}
Takeshi Kojima, Shixiang~Shane Gu, Machel Reid, Yutaka Matsuo, and Yusuke
  Iwasawa. 2022.
\newblock \href {https://openreview.net/forum?id=e2TBb5y0yFf} {Large language
  models are zero-shot reasoners}.
\newblock In \emph{Advances in Neural Information Processing Systems}.

\bibitem[{Lake and Baroni(2018)}]{lake2018generalization}
Brenden~M. Lake and Marco Baroni. 2018.
\newblock \href {http://proceedings.mlr.press/v80/lake18a.html} {Generalization
  without systematicity: On the compositional skills of sequence-to-sequence
  recurrent networks}.
\newblock In \emph{Proceedings of the 35th International Conference on Machine
  Learning, {ICML} 2018, Stockholmsm{\"{a}}ssan, Stockholm, Sweden, July 10-15,
  2018}, volume~80 of \emph{Proceedings of Machine Learning Research}, pages
  2879--2888. {PMLR}.

\bibitem[{Lampinen et~al.(2022)Lampinen, Dasgupta, Chan, Matthewson, Tessler,
  Creswell, McClelland, Wang, and Hill}]{lampinen2022can}
Andrew~K Lampinen, Ishita Dasgupta, Stephanie~CY Chan, Kory Matthewson,
  Michael~Henry Tessler, Antonia Creswell, James~L McClelland, Jane~X Wang, and
  Felix Hill. 2022.
\newblock Can language models learn from explanations in context?
\newblock In \emph{Findings of the Association for Computational Linguistics:
  EMNLP 2022}.

\bibitem[{Levi(2013)}]{levi2013introduction}
Edward~H Levi. 2013.
\newblock \emph{An introduction to legal reasoning}.
\newblock University of Chicago Press.

\bibitem[{Lewkowycz et~al.(2022)Lewkowycz, Andreassen, Dohan, Dyer,
  Michalewski, Ramasesh, Slone, Anil, Schlag, Gutman-Solo
  et~al.}]{lewkowycz2022solving}
Aitor Lewkowycz, Anders Andreassen, David Dohan, Ethan Dyer, Henryk
  Michalewski, Vinay Ramasesh, Ambrose Slone, Cem Anil, Imanol Schlag, Theo
  Gutman-Solo, et~al. 2022.
\newblock \href {https://arxiv.org/abs/2206.14858} {Solving quantitative
  reasoning problems with language models}.
\newblock \emph{ArXiv preprint}, abs/2206.14858.

\bibitem[{Li et~al.(2021)Li, Nye, and Andreas}]{li2021implicit}
Belinda~Z. Li, Maxwell Nye, and Jacob Andreas. 2021.
\newblock \href {https://doi.org/10.18653/v1/2021.acl-long.143} {Implicit
  representations of meaning in neural language models}.
\newblock In \emph{Proceedings of the 59th Annual Meeting of the Association
  for Computational Linguistics and the 11th International Joint Conference on
  Natural Language Processing (Volume 1: Long Papers)}, pages 1813--1827,
  Online. Association for Computational Linguistics.

\bibitem[{Li et~al.(2022{\natexlab{a}})Li, Chen, Shen, Chen, Zhang, Li, Wang,
  Qian, Peng, Mao et~al.}]{li2022explanations}
Shiyang Li, Jianshu Chen, Yelong Shen, Zhiyu Chen, Xinlu Zhang, Zekun Li, Hong
  Wang, Jing Qian, Baolin Peng, Yi~Mao, et~al. 2022{\natexlab{a}}.
\newblock \href {https://arxiv.org/abs/2210.06726} {Explanations from large
  language models make small reasoners better}.
\newblock \emph{ArXiv preprint}, abs/2210.06726.

\bibitem[{Li et~al.(2022{\natexlab{b}})Li, Lin, Zhang, Fu, Chen, Lou, and
  Chen}]{li2022advance}
Yifei Li, Zeqi Lin, Shizhuo Zhang, Qiang Fu, Bei Chen, Jian-Guang Lou, and
  Weizhu Chen. 2022{\natexlab{b}}.
\newblock \href {https://arxiv.org/abs/2206.02336} {On the advance of making
  language models better reasoners}.
\newblock \emph{ArXiv preprint}, abs/2206.02336.

\bibitem[{Lin et~al.(2020)Lin, Zhou, Shen, Zhou, Bhagavatula, Choi, and
  Ren}]{lin2020commongen}
Bill~Yuchen Lin, Wangchunshu Zhou, Ming Shen, Pei Zhou, Chandra Bhagavatula,
  Yejin Choi, and Xiang Ren. 2020.
\newblock \href {https://doi.org/10.18653/v1/2020.findings-emnlp.165}
  {{C}ommon{G}en: A constrained text generation challenge for generative
  commonsense reasoning}.
\newblock In \emph{Findings of the Association for Computational Linguistics:
  EMNLP 2020}, pages 1823--1840, Online. Association for Computational
  Linguistics.

\bibitem[{Ling et~al.(2017)Ling, Yogatama, Dyer, and Blunsom}]{ling2017program}
Wang Ling, Dani Yogatama, Chris Dyer, and Phil Blunsom. 2017.
\newblock \href {https://doi.org/10.18653/v1/P17-1015} {Program induction by
  rationale generation: Learning to solve and explain algebraic word problems}.
\newblock In \emph{Proceedings of the 55th Annual Meeting of the Association
  for Computational Linguistics (Volume 1: Long Papers)}, pages 158--167,
  Vancouver, Canada. Association for Computational Linguistics.

\bibitem[{Liu et~al.(2022{\natexlab{a}})Liu, Huang, Zhu, and
  Chang}]{liu2022dimongen}
Chenzhengyi Liu, Jie Huang, Kerui Zhu, and Kevin Chen-Chuan Chang.
  2022{\natexlab{a}}.
\newblock \href {https://arxiv.org/abs/2212.10545} {Dimongen: Diversified
  generative commonsense reasoning for explaining concept relationships}.
\newblock \emph{ArXiv preprint}, abs/2212.10545.

\bibitem[{Liu et~al.(2022{\natexlab{b}})Liu, Shen, Zhang, Dolan, Carin, and
  Chen}]{liu2022makes}
Jiachang Liu, Dinghan Shen, Yizhe Zhang, Bill Dolan, Lawrence Carin, and Weizhu
  Chen. 2022{\natexlab{b}}.
\newblock \href {https://doi.org/10.18653/v1/2022.deelio-1.10} {What makes good
  in-context examples for {GPT}-3?}
\newblock In \emph{Proceedings of Deep Learning Inside Out (DeeLIO 2022): The
  3rd Workshop on Knowledge Extraction and Integration for Deep Learning
  Architectures}, pages 100--114, Dublin, Ireland and Online. Association for
  Computational Linguistics.

\bibitem[{Liu et~al.(2019)Liu, Ott, Goyal, Du, Joshi, Chen, Levy, Lewis,
  Zettlemoyer, and Stoyanov}]{liu2019roberta}
Yinhan Liu, Myle Ott, Naman Goyal, Jingfei Du, Mandar Joshi, Danqi Chen, Omer
  Levy, Mike Lewis, Luke Zettlemoyer, and Veselin Stoyanov. 2019.
\newblock \href {https://arxiv.org/abs/1907.11692} {Roberta: A robustly
  optimized bert pretraining approach}.
\newblock \emph{ArXiv preprint}, abs/1907.11692.

\bibitem[{Lu et~al.(2022)Lu, Mishra, Xia, Qiu, Chang, Zhu, Tafjord, Clark, and
  Kalyan}]{lu2022learn}
Pan Lu, Swaroop Mishra, Tony Xia, Liang Qiu, Kai-Wei Chang, Song-Chun Zhu,
  Oyvind Tafjord, Peter Clark, and Ashwin Kalyan. 2022.
\newblock \href {https://openreview.net/forum?id=HjwK-Tc_Bc} {Learn to explain:
  Multimodal reasoning via thought chains for science question answering}.
\newblock In \emph{Advances in Neural Information Processing Systems}.

\bibitem[{Madaan et~al.(2022)Madaan, Zhou, Alon, Yang, and
  Neubig}]{madaan2022language}
Aman Madaan, Shuyan Zhou, Uri Alon, Yiming Yang, and Graham Neubig. 2022.
\newblock Language models of code are few-shot commonsense learners.
\newblock In \emph{Proceedings of the 2022 Conference on Empirical Methods in
  Natural Language Processing (EMNLP)}.

\bibitem[{Magister et~al.(2022)Magister, Mallinson, Adamek, Malmi, and
  Severyn}]{magister2022@teaching}
Lucie~Charlotte Magister, Jonathan Mallinson, Jakub Adamek, Eric Malmi, and
  Aliaksei Severyn. 2022.
\newblock \href {https://arxiv.org/abs/2212.08410} {Teaching small language
  models to reason}.
\newblock \emph{ArXiv preprint}, abs/2212.08410.

\bibitem[{Manning(2022)}]{manning2022human}
Christopher~D Manning. 2022.
\newblock Human language understanding \& reasoning.
\newblock \emph{Daedalus}, 151(2):127--138.

\bibitem[{Marcus(2020)}]{marcus2020next}
Gary Marcus. 2020.
\newblock \href {https://arxiv.org/abs/2002.06177} {The next decade in ai: four
  steps towards robust artificial intelligence}.
\newblock \emph{ArXiv preprint}, abs/2002.06177.

\bibitem[{McHugh and Way(2018)}]{mchugh2018reasoning}
Conor McHugh and Jonathan Way. 2018.
\newblock What is reasoning?
\newblock \emph{Mind}, 127(505):167--196.

\bibitem[{Miao et~al.(2020)Miao, Liang, and Su}]{miao2020diverse}
Shen-yun Miao, Chao-Chun Liang, and Keh-Yih Su. 2020.
\newblock \href {https://doi.org/10.18653/v1/2020.acl-main.92} {A diverse
  corpus for evaluating and developing {E}nglish math word problem solvers}.
\newblock In \emph{Proceedings of the 58th Annual Meeting of the Association
  for Computational Linguistics}, pages 975--984, Online. Association for
  Computational Linguistics.

\bibitem[{Min et~al.(2019)Min, Zhong, Zettlemoyer, and
  Hajishirzi}]{min2019multi}
Sewon Min, Victor Zhong, Luke Zettlemoyer, and Hannaneh Hajishirzi. 2019.
\newblock \href {https://doi.org/10.18653/v1/P19-1613} {Multi-hop reading
  comprehension through question decomposition and rescoring}.
\newblock In \emph{Proceedings of the 57th Annual Meeting of the Association
  for Computational Linguistics}, pages 6097--6109, Florence, Italy.
  Association for Computational Linguistics.

\bibitem[{Misra et~al.(2022)Misra, Rayz, and Ettinger}]{misra2022property}
Kanishka Misra, Julia~Taylor Rayz, and Allyson Ettinger. 2022.
\newblock \href {https://arxiv.org/abs/2205.06910} {A property induction
  framework for neural language models}.
\newblock \emph{ArXiv preprint}, abs/2205.06910.

\bibitem[{Mitchell(2021)}]{mitchell2021abstraction}
Melanie Mitchell. 2021.
\newblock Abstraction and analogy-making in artificial intelligence.
\newblock \emph{Annals of the New York Academy of Sciences}, 1505(1):79--101.

\bibitem[{Nan et~al.(2022)Nan, Hsieh, Mao, Lin, Verma, Zhang,
  Kry{\'s}ci{\'n}ski, Schoelkopf, Kong, Tang, Mutuma, Rosand, Trindade,
  Bandaru, Cunningham, Xiong, Radev, and Radev}]{nan2022fetaqa}
Linyong Nan, Chiachun Hsieh, Ziming Mao, Xi~Victoria Lin, Neha Verma, Rui
  Zhang, Wojciech Kry{\'s}ci{\'n}ski, Hailey Schoelkopf, Riley Kong, Xiangru
  Tang, Mutethia Mutuma, Ben Rosand, Isabel Trindade, Renusree Bandaru, Jacob
  Cunningham, Caiming Xiong, Dragomir Radev, and Dragomir Radev. 2022.
\newblock \href {https://doi.org/10.1162/tacl_a_00446} {{F}e{T}a{QA}: Free-form
  table question answering}.
\newblock \emph{Transactions of the Association for Computational Linguistics},
  10:35--49.

\bibitem[{Nye et~al.(2022)Nye, Andreassen, Gur-Ari, Michalewski, Austin,
  Bieber, Dohan, Lewkowycz, Bosma, Luan, Sutton, and Odena}]{nye2021show}
Maxwell Nye, Anders~Johan Andreassen, Guy Gur-Ari, Henryk Michalewski, Jacob
  Austin, David Bieber, David Dohan, Aitor Lewkowycz, Maarten Bosma, David
  Luan, Charles Sutton, and Augustus Odena. 2022.
\newblock \href {https://openreview.net/forum?id=HBlx2idbkbq} {Show your work:
  Scratchpads for intermediate computation with language models}.
\newblock In \emph{Deep Learning for Code Workshop}.

\bibitem[{OpenAI(2022)}]{openai2022chatgpt}
OpenAI. 2022.
\newblock Chatgpt: Optimizing language models for dialogue.
\newblock \emph{OpenAI}.

\bibitem[{Passmore(1961)}]{passmore1961philosophical}
John~Arthur Passmore. 1961.
\newblock Philosophical reasoning.

\bibitem[{Pasupat and Liang(2015)}]{pasupat2015compositional}
Panupong Pasupat and Percy Liang. 2015.
\newblock \href {https://doi.org/10.3115/v1/P15-1142} {Compositional semantic
  parsing on semi-structured tables}.
\newblock In \emph{Proceedings of the 53rd Annual Meeting of the Association
  for Computational Linguistics and the 7th International Joint Conference on
  Natural Language Processing (Volume 1: Long Papers)}, pages 1470--1480,
  Beijing, China. Association for Computational Linguistics.

\bibitem[{Patel et~al.(2021)Patel, Bhattamishra, and Goyal}]{patel2021nlp}
Arkil Patel, Satwik Bhattamishra, and Navin Goyal. 2021.
\newblock \href {https://doi.org/10.18653/v1/2021.naacl-main.168} {Are {NLP}
  models really able to solve simple math word problems?}
\newblock In \emph{Proceedings of the 2021 Conference of the North American
  Chapter of the Association for Computational Linguistics: Human Language
  Technologies}, pages 2080--2094, Online. Association for Computational
  Linguistics.

\bibitem[{Perez et~al.(2020)Perez, Lewis, Yih, Cho, and
  Kiela}]{perez2020unsupervised}
Ethan Perez, Patrick Lewis, Wen-tau Yih, Kyunghyun Cho, and Douwe Kiela. 2020.
\newblock \href {https://doi.org/10.18653/v1/2020.emnlp-main.713} {Unsupervised
  question decomposition for question answering}.
\newblock In \emph{Proceedings of the 2020 Conference on Empirical Methods in
  Natural Language Processing (EMNLP)}, pages 8864--8880, Online. Association
  for Computational Linguistics.

\bibitem[{Pi et~al.(2022)Pi, Liu, Chen, Ziyadi, Lin, Gao, Fu, Lou, and
  Chen}]{pi2022reasoning}
Xinyu Pi, Qian Liu, Bei Chen, Morteza Ziyadi, Zeqi Lin, Yan Gao, Qiang Fu,
  Jian-Guang Lou, and Weizhu Chen. 2022.
\newblock Reasoning like program executors.
\newblock In \emph{Proceedings of the 2022 Conference on Empirical Methods in
  Natural Language Processing (EMNLP)}.

\bibitem[{Piantasodi and Hill(2022)}]{piantasodi2022meaning}
Steven~T Piantasodi and Felix Hill. 2022.
\newblock \href {https://arxiv.org/abs/2208.02957} {Meaning without reference
  in large language models}.
\newblock \emph{ArXiv preprint}, abs/2208.02957.

\bibitem[{Press et~al.(2022)Press, Zhang, Min, Schmidt, Smith, and
  Lewis}]{press2022measuring}
Ofir Press, Muru Zhang, Sewon Min, Ludwig Schmidt, Noah~A Smith, and Mike
  Lewis. 2022.
\newblock \href {https://arxiv.org/abs/2210.03350} {Measuring and narrowing the
  compositionality gap in language models}.
\newblock \emph{ArXiv preprint}, abs/2210.03350.

\bibitem[{Prystawski et~al.(2022)Prystawski, Thibodeau, and
  Goodman}]{prystawski2022psychologically}
Ben Prystawski, Paul Thibodeau, and Noah Goodman. 2022.
\newblock \href {https://arxiv.org/abs/2209.08141} {Psychologically-informed
  chain-of-thought prompts for metaphor understanding in large language
  models}.
\newblock \emph{ArXiv preprint}, abs/2209.08141.

\bibitem[{Qiao et~al.(2022)Qiao, Ou, Zhang, Chen, Yao, Deng, Tan, Huang, and
  Chen}]{qiao2022reasoning}
Shuofei Qiao, Yixin Ou, Ningyu Zhang, Xiang Chen, Yunzhi Yao, Shumin Deng,
  Chuanqi Tan, Fei Huang, and Huajun Chen. 2022.
\newblock \href {https://arxiv.org/abs/2212.09597} {Reasoning with language
  model prompting: A survey}.
\newblock \emph{ArXiv preprint}, abs/2212.09597.

\bibitem[{Radford et~al.(2018)Radford, Narasimhan, Salimans, Sutskever
  et~al.}]{radford2018improving}
Alec Radford, Karthik Narasimhan, Tim Salimans, Ilya Sutskever, et~al. 2018.
\newblock Improving language understanding by generative pre-training.

\bibitem[{Radford et~al.(2019)Radford, Wu, Child, Luan, Amodei, Sutskever
  et~al.}]{radford2019language}
Alec Radford, Jeffrey Wu, Rewon Child, David Luan, Dario Amodei, Ilya
  Sutskever, et~al. 2019.
\newblock Language models are unsupervised multitask learners.
\newblock \emph{OpenAI blog}, 1(8):9.

\bibitem[{Rae et~al.(2021)Rae, Borgeaud, Cai, Millican, Hoffmann, Song,
  Aslanides, Henderson, Ring, Young et~al.}]{rae2021scaling}
Jack~W Rae, Sebastian Borgeaud, Trevor Cai, Katie Millican, Jordan Hoffmann,
  Francis Song, John Aslanides, Sarah Henderson, Roman Ring, Susannah Young,
  et~al. 2021.
\newblock \href {https://arxiv.org/abs/2112.11446} {Scaling language models:
  Methods, analysis \& insights from training gopher}.
\newblock \emph{ArXiv preprint}, abs/2112.11446.

\bibitem[{Raffel et~al.(2020)Raffel, Shazeer, Roberts, Lee, Narang, Matena,
  Zhou, Li, Liu et~al.}]{raffel2020exploring}
Colin Raffel, Noam Shazeer, Adam Roberts, Katherine Lee, Sharan Narang, Michael
  Matena, Yanqi Zhou, Wei Li, Peter~J Liu, et~al. 2020.
\newblock Exploring the limits of transfer learning with a unified text-to-text
  transformer.
\newblock \emph{J. Mach. Learn. Res.}, 21(140):1--67.

\bibitem[{Rajani et~al.(2019)Rajani, McCann, Xiong, and
  Socher}]{rajani2019explain}
Nazneen~Fatema Rajani, Bryan McCann, Caiming Xiong, and Richard Socher. 2019.
\newblock \href {https://doi.org/10.18653/v1/P19-1487} {Explain yourself!
  leveraging language models for commonsense reasoning}.
\newblock In \emph{Proceedings of the 57th Annual Meeting of the Association
  for Computational Linguistics}, pages 4932--4942, Florence, Italy.
  Association for Computational Linguistics.

\bibitem[{Razeghi et~al.(2022)Razeghi, Logan~IV, Gardner, and
  Singh}]{razeghi2022impact}
Yasaman Razeghi, Robert~L Logan~IV, Matt Gardner, and Sameer Singh. 2022.
\newblock \href {https://arxiv.org/abs/2202.07206} {Impact of pretraining term
  frequencies on few-shot reasoning}.
\newblock \emph{ArXiv preprint}, abs/2202.07206.

\bibitem[{Roy and Roth(2015)}]{roy2015solving}
Subhro Roy and Dan Roth. 2015.
\newblock \href {https://doi.org/10.18653/v1/D15-1202} {Solving general
  arithmetic word problems}.
\newblock In \emph{Proceedings of the 2015 Conference on Empirical Methods in
  Natural Language Processing}, pages 1743--1752, Lisbon, Portugal. Association
  for Computational Linguistics.

\bibitem[{Ruis et~al.(2022)Ruis, Khan, Biderman, Hooker, Rockt{\"a}schel, and
  Grefenstette}]{ruis2022large}
Laura Ruis, Akbir Khan, Stella Biderman, Sara Hooker, Tim Rockt{\"a}schel, and
  Edward Grefenstette. 2022.
\newblock \href {https://arxiv.org/abs/2210.14986} {Large language models are
  not zero-shot communicators}.
\newblock \emph{ArXiv preprint}, abs/2210.14986.

\bibitem[{Russin et~al.(2020)Russin, O’Reilly, and Bengio}]{russin2020deep}
Jacob Russin, Randall~C O’Reilly, and Yoshua Bengio. 2020.
\newblock Deep learning needs a prefrontal cortex.
\newblock \emph{Work Bridging AI Cogn Sci}, 107:603--616.

\bibitem[{Saparov and He(2022)}]{saparov2022language}
Abulhair Saparov and He~He. 2022.
\newblock \href {https://arxiv.org/abs/2210.01240} {Language models are greedy
  reasoners: A systematic formal analysis of chain-of-thought}.
\newblock \emph{ArXiv preprint}, abs/2210.01240.

\bibitem[{Scao et~al.(2022)Scao, Fan, Akiki, Pavlick, Ili{\'c}, Hesslow,
  Castagn{\'e}, Luccioni, Yvon, Gall{\'e} et~al.}]{scao2022bloom}
Teven~Le Scao, Angela Fan, Christopher Akiki, Ellie Pavlick, Suzana Ili{\'c},
  Daniel Hesslow, Roman Castagn{\'e}, Alexandra~Sasha Luccioni, Fran{\c{c}}ois
  Yvon, Matthias Gall{\'e}, et~al. 2022.
\newblock \href {https://arxiv.org/abs/2211.05100} {Bloom: A 176b-parameter
  open-access multilingual language model}.
\newblock \emph{ArXiv preprint}, abs/2211.05100.

\bibitem[{Shi et~al.(2022)Shi, Suzgun, Freitag, Wang, Srivats, Vosoughi, Chung,
  Tay, Ruder, Zhou et~al.}]{shi2022language}
Freda Shi, Mirac Suzgun, Markus Freitag, Xuezhi Wang, Suraj Srivats, Soroush
  Vosoughi, Hyung~Won Chung, Yi~Tay, Sebastian Ruder, Denny Zhou, et~al. 2022.
\newblock \href {https://arxiv.org/abs/2210.03057} {Language models are
  multilingual chain-of-thought reasoners}.
\newblock \emph{ArXiv preprint}, abs/2210.03057.

\bibitem[{Shridhar et~al.(2022)Shridhar, Stolfo, and
  Sachan}]{shridhar2022distilling}
Kumar Shridhar, Alessandro Stolfo, and Mrinmaya Sachan. 2022.
\newblock \href {https://arxiv.org/abs/2212.00193} {Distilling multi-step
  reasoning capabilities of large language models into smaller models via
  semantic decompositions}.
\newblock \emph{ArXiv preprint}, abs/2212.00193.

\bibitem[{Song et~al.(2022)Song, Wu, Washington, Sadler, Chao, and
  Su}]{song2022llm}
Chan~Hee Song, Jiaman Wu, Clayton Washington, Brian~M Sadler, Wei-Lun Chao, and
  Yu~Su. 2022.
\newblock \href {https://arxiv.org/abs/2212.04088} {Llm-planner: Few-shot
  grounded planning for embodied agents with large language models}.
\newblock \emph{ArXiv preprint}, abs/2212.04088.

\bibitem[{Srivastava et~al.(2022)Srivastava, Rastogi, Rao, Shoeb, Abid, Fisch,
  Brown, Santoro, Gupta, Garriga-Alonso et~al.}]{srivastava2022beyond}
Aarohi Srivastava, Abhinav Rastogi, Abhishek Rao, Abu Awal~Md Shoeb, Abubakar
  Abid, Adam Fisch, Adam~R Brown, Adam Santoro, Aditya Gupta, Adri{\`a}
  Garriga-Alonso, et~al. 2022.
\newblock \href {https://arxiv.org/abs/2206.04615} {Beyond the imitation game:
  Quantifying and extrapolating the capabilities of language models}.
\newblock \emph{ArXiv preprint}, abs/2206.04615.

\bibitem[{Suzgun et~al.(2022)Suzgun, Scales, Sch{\"a}rli, Gehrmann, Tay, Chung,
  Chowdhery, Le, Chi, Zhou et~al.}]{suzgun2022challenging}
Mirac Suzgun, Nathan Scales, Nathanael Sch{\"a}rli, Sebastian Gehrmann, Yi~Tay,
  Hyung~Won Chung, Aakanksha Chowdhery, Quoc~V Le, Ed~H Chi, Denny Zhou, et~al.
  2022.
\newblock \href {https://arxiv.org/abs/2210.09261} {Challenging big-bench tasks
  and whether chain-of-thought can solve them}.
\newblock \emph{ArXiv preprint}, abs/2210.09261.

\bibitem[{Talmor and Berant(2018)}]{talmor2018web}
Alon Talmor and Jonathan Berant. 2018.
\newblock \href {https://doi.org/10.18653/v1/N18-1059} {The web as a
  knowledge-base for answering complex questions}.
\newblock In \emph{Proceedings of the 2018 Conference of the North {A}merican
  Chapter of the Association for Computational Linguistics: Human Language
  Technologies, Volume 1 (Long Papers)}, pages 641--651, New Orleans,
  Louisiana. Association for Computational Linguistics.

\bibitem[{Talmor et~al.(2019)Talmor, Herzig, Lourie, and
  Berant}]{talmor2019commonsenseqa}
Alon Talmor, Jonathan Herzig, Nicholas Lourie, and Jonathan Berant. 2019.
\newblock \href {https://doi.org/10.18653/v1/N19-1421} {{C}ommonsense{QA}: A
  question answering challenge targeting commonsense knowledge}.
\newblock In \emph{Proceedings of the 2019 Conference of the North {A}merican
  Chapter of the Association for Computational Linguistics: Human Language
  Technologies, Volume 1 (Long and Short Papers)}, pages 4149--4158,
  Minneapolis, Minnesota. Association for Computational Linguistics.

\bibitem[{Talmor et~al.(2020)Talmor, Tafjord, Clark, Goldberg, and
  Berant}]{talmor2020leap}
Alon Talmor, Oyvind Tafjord, Peter Clark, Yoav Goldberg, and Jonathan Berant.
  2020.
\newblock \href
  {https://proceedings.neurips.cc/paper/2020/hash/e992111e4ab9985366e806733383bd8c-Abstract.html}
  {Leap-of-thought: Teaching pre-trained models to systematically reason over
  implicit knowledge}.
\newblock In \emph{Advances in Neural Information Processing Systems 33: Annual
  Conference on Neural Information Processing Systems 2020, NeurIPS 2020,
  December 6-12, 2020, virtual}.

\bibitem[{Taylor et~al.(2022)Taylor, Kardas, Cucurull, Scialom, Hartshorn,
  Saravia, Poulton, Kerkez, and Stojnic}]{taylor2022galactica}
Ross Taylor, Marcin Kardas, Guillem Cucurull, Thomas Scialom, Anthony
  Hartshorn, Elvis Saravia, Andrew Poulton, Viktor Kerkez, and Robert Stojnic.
  2022.
\newblock \href {https://arxiv.org/abs/2211.09085} {Galactica: A large language
  model for science}.
\newblock \emph{ArXiv preprint}, abs/2211.09085.

\bibitem[{Valmeekam et~al.(2022)Valmeekam, Olmo, Sreedharan, and
  Kambhampati}]{valmeekam2022large}
Karthik Valmeekam, Alberto Olmo, Sarath Sreedharan, and Subbarao Kambhampati.
  2022.
\newblock Large language models still can't plan (a benchmark for llms on
  planning and reasoning about change).
\newblock In \emph{NeurIPS 2022 Foundation Models for Decision Making
  Workshop}.

\bibitem[{Wang et~al.(2022{\natexlab{a}})Wang, Deng, and
  Sun}]{Wang2022iteratively}
Boshi Wang, Xiang Deng, and Huan Sun. 2022{\natexlab{a}}.
\newblock Iteratively prompt pre-trained language models for chain of thought.
\newblock In \emph{The 2022 Conference on Empirical Methods for Natural
  Language Processing}.

\bibitem[{Wang et~al.(2022{\natexlab{b}})Wang, Min, Deng, Shen, Wu,
  Zettlemoyer, and Sun}]{wang2022towards}
Boshi Wang, Sewon Min, Xiang Deng, Jiaming Shen, You Wu, Luke Zettlemoyer, and
  Huan Sun. 2022{\natexlab{b}}.
\newblock \href {https://arxiv.org/abs/2212.10001} {Towards understanding
  chain-of-thought prompting: An empirical study of what matters}.
\newblock \emph{ArXiv preprint}, abs/2212.10001.

\bibitem[{Wang et~al.(2022{\natexlab{c}})Wang, Wei, Schuurmans, Le, Chi, and
  Zhou}]{wang2022self}
Xuezhi Wang, Jason Wei, Dale Schuurmans, Quoc Le, Ed~Chi, and Denny Zhou.
  2022{\natexlab{c}}.
\newblock \href {https://arxiv.org/abs/2203.11171} {Self-consistency improves
  chain of thought reasoning in language models}.
\newblock \emph{ArXiv preprint}, abs/2203.11171.

\bibitem[{Wason(1968)}]{wason1968reasoning}
Peter~C Wason. 1968.
\newblock Reasoning about a rule.
\newblock \emph{Quarterly journal of experimental psychology}, 20(3):273--281.

\bibitem[{Wason and Johnson-Laird(1972)}]{wason1972psychology}
Peter~Cathcart Wason and Philip~Nicholas Johnson-Laird. 1972.
\newblock \emph{Psychology of reasoning: Structure and content}, volume~86.
\newblock Harvard University Press.

\bibitem[{Wei et~al.(2022{\natexlab{a}})Wei, Tay, Bommasani, Raffel, Zoph,
  Borgeaud, Yogatama, Bosma, Zhou, Metzler et~al.}]{wei2022emergent}
Jason Wei, Yi~Tay, Rishi Bommasani, Colin Raffel, Barret Zoph, Sebastian
  Borgeaud, Dani Yogatama, Maarten Bosma, Denny Zhou, Donald Metzler, et~al.
  2022{\natexlab{a}}.
\newblock Emergent abilities of large language models.
\newblock \emph{Transactions on Machine Learning Research}.

\bibitem[{Wei et~al.(2022{\natexlab{b}})Wei, Wang, Schuurmans, Bosma, brian
  ichter, Xia, Chi, Le, and Zhou}]{wei2022chain}
Jason Wei, Xuezhi Wang, Dale Schuurmans, Maarten Bosma, brian ichter, Fei Xia,
  Ed~H. Chi, Quoc~V Le, and Denny Zhou. 2022{\natexlab{b}}.
\newblock \href {https://openreview.net/forum?id=_VjQlMeSB_J} {Chain of thought
  prompting elicits reasoning in large language models}.
\newblock In \emph{Advances in Neural Information Processing Systems}.

\bibitem[{Weng et~al.(2022)Weng, Zhu, He, Liu, and Zhao}]{weng2022large}
Yixuan Weng, Minjun Zhu, Shizhu He, Kang Liu, and Jun Zhao. 2022.
\newblock \href {https://arxiv.org/abs/2212.09561} {Large language models are
  reasoners with self-verification}.
\newblock \emph{ArXiv preprint}, abs/2212.09561.

\bibitem[{Wiegreffe et~al.(2022)Wiegreffe, Hessel, Swayamdipta, Riedl, and
  Choi}]{wiegreffe2021reframing}
Sarah Wiegreffe, Jack Hessel, Swabha Swayamdipta, Mark Riedl, and Yejin Choi.
  2022.
\newblock \href {https://doi.org/10.18653/v1/2022.naacl-main.47} {Reframing
  human-{AI} collaboration for generating free-text explanations}.
\newblock In \emph{Proceedings of the 2022 Conference of the North American
  Chapter of the Association for Computational Linguistics: Human Language
  Technologies}, pages 632--658, Seattle, United States. Association for
  Computational Linguistics.

\bibitem[{Yang et~al.(2022)Yang, Dong, Du, Cheng, Cambria, Liu, Gao, and
  Wei}]{yang2022language}
Zonglin Yang, Li~Dong, Xinya Du, Hao Cheng, Erik Cambria, Xiaodong Liu,
  Jianfeng Gao, and Furu Wei. 2022.
\newblock \href {https://arxiv.org/abs/2212.10923} {Language models as
  inductive reasoners}.
\newblock \emph{ArXiv preprint}, abs/2212.10923.

\bibitem[{Ye and Durrett(2022)}]{ye2022unreliability}
Xi~Ye and Greg Durrett. 2022.
\newblock The unreliability of explanations in few-shot prompting for textual
  reasoning.
\newblock \emph{Advances in neural information processing systems}.

\bibitem[{Yu et~al.(2022)Yu, Wang, Golovneva, Alkhamissy, Ghosh, Diab, and
  Celikyilmaz}]{yu2022alert}
Ping Yu, Tianlu Wang, Olga Golovneva, Badr Alkhamissy, Gargi Ghosh, Mona Diab,
  and Asli Celikyilmaz. 2022.
\newblock \href {https://arxiv.org/abs/2212.08286} {Alert: Adapting language
  models to reasoning tasks}.
\newblock \emph{ArXiv preprint}, abs/2212.08286.

\bibitem[{Zelikman et~al.(2022)Zelikman, Wu, Mu, and
  Goodman}]{zelikman2022star}
Eric Zelikman, Yuhuai Wu, Jesse Mu, and Noah Goodman. 2022.
\newblock \href {https://openreview.net/forum?id=_3ELRdg2sgI} {{ST}ar:
  Bootstrapping reasoning with reasoning}.
\newblock In \emph{Advances in Neural Information Processing Systems}.

\bibitem[{Zhang et~al.(2022{\natexlab{a}})Zhang, Roller, Goyal, Artetxe, Chen,
  Chen, Dewan, Diab, Li, Lin et~al.}]{zhang2022opt}
Susan Zhang, Stephen Roller, Naman Goyal, Mikel Artetxe, Moya Chen, Shuohui
  Chen, Christopher Dewan, Mona Diab, Xian Li, Xi~Victoria Lin, et~al.
  2022{\natexlab{a}}.
\newblock \href {https://arxiv.org/abs/2205.01068} {Opt: Open pre-trained
  transformer language models}.
\newblock \emph{ArXiv preprint}, abs/2205.01068.

\bibitem[{Zhang et~al.(2022{\natexlab{b}})Zhang, Zhang, Li, and
  Smola}]{zhang2022automatic}
Zhuosheng Zhang, Aston Zhang, Mu~Li, and Alex Smola. 2022{\natexlab{b}}.
\newblock \href {https://arxiv.org/abs/2210.03493} {Automatic chain of thought
  prompting in large language models}.
\newblock \emph{ArXiv preprint}, abs/2210.03493.

\bibitem[{Zheng et~al.(2023)Zheng, Huang, and Chang}]{zheng2023does}
Shen Zheng, Jie Huang, and Kevin Chen-Chuan Chang. 2023.
\newblock \href {https://arxiv.org/abs/2304.10513} {Why does chatgpt fall short
  in providing truthful answers?}
\newblock \emph{ArXiv preprint}, abs/2304.10513.

\bibitem[{Zhou et~al.(2022{\natexlab{a}})Zhou, Sch{\"a}rli, Hou, Wei, Scales,
  Wang, Schuurmans, Bousquet, Le, and Chi}]{zhou2022least}
Denny Zhou, Nathanael Sch{\"a}rli, Le~Hou, Jason Wei, Nathan Scales, Xuezhi
  Wang, Dale Schuurmans, Olivier Bousquet, Quoc Le, and Ed~Chi.
  2022{\natexlab{a}}.
\newblock \href {https://arxiv.org/abs/2205.10625} {Least-to-most prompting
  enables complex reasoning in large language models}.
\newblock \emph{ArXiv preprint}, abs/2205.10625.

\bibitem[{Zhou et~al.(2022{\natexlab{b}})Zhou, Dong, Liu, Cheng, Han, and
  Zhang}]{zhou2022reflection}
Fan Zhou, Haoyu Dong, Qian Liu, Zhoujun Cheng, Shi Han, and Dongmei Zhang.
  2022{\natexlab{b}}.
\newblock \href {https://arxiv.org/abs/2210.05075} {Reflection of thought:
  Inversely eliciting numerical reasoning in language models via solving linear
  systems}.
\newblock \emph{ArXiv preprint}, abs/2210.05075.

\bibitem[{Zhou et~al.(2022{\natexlab{c}})Zhou, Nova, Larochelle, Courville,
  Neyshabur, and Sedghi}]{zhou2022teaching}
Hattie Zhou, Azade Nova, Hugo Larochelle, Aaron Courville, Behnam Neyshabur,
  and Hanie Sedghi. 2022{\natexlab{c}}.
\newblock \href {https://arxiv.org/abs/2211.09066} {Teaching algorithmic
  reasoning via in-context learning}.
\newblock \emph{ArXiv preprint}, abs/2211.09066.

\bibitem[{Zimmerman(2000)}]{zimmerman2000development}
Corinne Zimmerman. 2000.
\newblock The development of scientific reasoning skills.
\newblock \emph{Developmental review}, 20(1):99--149.

\end{thebibliography}
\bibliographystyle{acl_natbib}

\end{document}